\documentclass[12pt]{spieman}  % 12pt font required by SPIE;
\usepackage{amsmath,amsfonts,amssymb}
\usepackage{pythonhighlight}
\usepackage{graphicx}
\usepackage{setspace}
\usepackage{tocloft}
\usepackage{multirow}
\usepackage{xcolor}
\usepackage{lineno}

\usepackage[capitalize]{cleveref}

\title{Margin-Aware Intra-Class Novelty Identification for Medical Images}

\author[a,*]{Xiaoyuan Guo}
\author[b,c]{Judy Wawira Gichoya}
\author[d]{Saptarshi Purkayastha}
\author[e]{Imon Banerjee}
\affil[a]{Emory University, Department of Computer Science}
\affil[b]{Emory University, Department of Radiology and Imaging Sciences}
\affil[c]{Emory University, Department of Biomedical Informatics}
\affil[d]{Indiana University-Purdue University Indianapolis, School of Informatics and Computing}
\affil[e]{Mayo clinic, Arizona State University}

\cftpagenumbersoff{figure}
\cftpagenumbersoff{table} 
\begin{document} 
\maketitle

\begin{abstract}

\noindent {\bf Purpose:}
Existing anomaly detection methods focus on detecting inter-class variations while medical image novelty identification is more challenging in the presence of intra-class variations. For example, a model trained with normal chest X-ray and common lung abnormalities, is expected to discover and flag idiopathic pulmonary fibrosis which is a rare lung disease and unseen during training. The nuances of intra-class variations and lack of relevant training data in medical image analysis pose great challenges for existing anomaly detection methods.

\noindent {\bf Approach:}
We address the above challenges by proposing a hybrid model - \textbf{T}ransformation-based \textbf{E}mbedding learning for \textbf{N}ovelty \textbf{D}etection (\textbf{TEND}), which combines the merits of Classifier-based approach and AutoEncoder based approach. Training TEND consists of two stages. In the first stage, we learn in-distribution embeddings with an AutoEncoder via the unsupervised reconstruction. In the second stage, we learn a discriminative classifier to distinguish in-distribution data and the transformed counterparts. Additionally, we propose a margin-aware objective to pull in-distribution data in a hypersphere while pushing away the transformed data. Eventually, the weighted sum of class probability and the distance to margin constitutes the anomaly score. 

\noindent {\bf Results:}
Extensive experiments are performed on three public medical image datasets with the one-vs-rest setup (namely one class as in-distribution data and the left as intra-class out-of-distribution data) and the rest-vs-one setup. Additional experiments on generated intra-class out-of-distribution data with unused transformations are implemented on the datasets. The quantitative results show competitive performance as compared to the state-of-the-art approaches. Provided qualitative examples further demonstrate the effectiveness of TEND.   

\noindent {\bf Conclusion:} 
Our anomaly detection model TEND can effectively identify the challenging intra-class out-of-distribution medical images in an unsupervised fashion. It can be applied to discover unseen medical image classes and serve as the abnormal data screening for downstream medical tasks. The corresponding code is available at \url{https://github.com/XiaoyuanGuo/TEND_MedicalNoveltyDetection}. 
\end{abstract}

% Include a list of up to six keywords after the abstract

% Include email contact information for corresponding author
% {\noindent \footnotesize\textbf{*}Imon Banerjee,  \linkable{Banerjee.Imon@mayo.edu} }
{\noindent \footnotesize\textbf{*}Xiaoyuan Guo,  \linkable{xiaoyuan.guo@emory.edu} }

\begin{spacing}{2}   % use double spacing for rest of manuscript

\section{Introduction}
\label{sect:intro}  % \label{} allows reference to this section

%Add distribution shifting plots
With recent prominent developments of machine learning techniques in computer vision, integrating machine learning tools to solve medical image problems is becoming more and more popular due to the powerful computation and efficiency~\cite{litjens2017survey}. However, when deploying machine learning models in real-world applications, models trained on in-distribution (ID) data may fail to deal with out-of-distribution (OOD) inputs and assign incorrect probabilities~\cite{song2020critical}. This can severely contaminate the reliability of artificial intelligence models, especially in medical areas as the safety in clinical decisions is much more critical than other fields. For example, a classifier trained on existing bacterial classes wrongly classified a new type of bacteria as one of the classes from the training data with high confidence~\cite{ren2019likelihood:likelihood}, which could be concerning for clinical usage but may be avoided by combining an OOD detection model. Thus, a successful open-world deployment with OOD detection should be sensitive to unseen classes and distribution-shifted samples and also be resilient to potential adversarial attacks~\cite{sehwag2019analyzing}.  

However, medical OOD detection poses great challenges due to the heterogeneity and unknown data characteristics of medical data. 1) \emph{Mutations can happen.} Different from natural objects with fixed attributes, known diseases may progress to other mutated versions and generate anomalous data; 
% New patterns of mutations are being discovered every year.
2) \emph{Heterogeneous data is a big concern.} Medical images collected from different race groups can introduce heterogeneity; 3) \emph{Distribution shifting always exists.} Data scanned with different machines or institutes may have distribution shifting; 4) \emph{Data with defects is common.} Medical images can be overexposed or scanned with incorrect positions/angles.

\textit{OOD} data, also called \textit{anomaly}, \textit{outlier}, usually refers to data that shows dissimilarity from the training distribution. Given an image $x$, the goal of \textit{OOD detection} is to identify whether $x$ is from ID dataset $D_{in}$ or OOD dataset $D_{out}$. 
There are two types of OOD data commonly targeted to identify - \textit{(i) intra-class data:} OOD data belonging this type, which is also called \textit{novelty data}, often shares severe similarity with the ID classes and is extremely challenging to distinguish, \textit{e.g.}, the pneumonia chest X-ray presents close appearance with the normal images; \textit{(ii) inter-class data:} this data is significantly different from ID samples, \textit{e.g.,} a head CT image is much different in shape and color from the skin cancer image. Even though many anomaly detection methods have been proposed~\cite{schlachter2019deep:schlachter,liznerski2020explainable:explainable}, most of them focus on natural images and follow the one-vs-rest setup~\cite{tack2020csi:tack} for benchmark natural image datasets (\textit{e.g.,} MNIST~\cite{lecun-mnisthandwrittendigit-2010}, Fashion-MNIST~\cite{xiao2017fashion}, CIFAR-10~\cite{krizhevsky2010cifar}, ImageNet~\cite{deng2009imagenet}, \textit{etc.}). Thus, the performance reported on the benchmark datasets is actually for inter-class prediction due to the clear class variation and often trivial to detect. In contrast, the anomaly detection in medical images is more of an \emph{intra-class} identification problem, which can be also called \textit{novelty detection}.

% 2. Provide key insights -- how to solve the key challenge mentioned above. 
% (1) why reconstruction--finer details, learn transformations. 
% (2) why metric learning
% (3) classification is used to learn clusters
To train a novelty detector with only ID data available, learning high-quality ``normality" features is the fundamental step to identify the OOD samples during inference. AutoEncoder~\cite{mcclelland1986parallel} architecture, as an unsupervised model to learn efficient data features through reconstruction, is the most straightforward way to extract features for ID data~\cite{ruff2018deep:ruff}. For anomaly detection, the reconstruction error is treated as the score of outliers based on the assumption that the AutoEncoder~\cite{mcclelland1986parallel} is unable to reconstruct the anomalies well and causes large reconstruction errors. However, in the intra-class detection where the variations among the in-class and out-of-class medical images of the same category are very subtle, the AutoEncoder~\cite{mcclelland1986parallel} often fails owing to the lack of discriminative ability for intra-class detection (see Sec.~\ref{sec:background}).

% 3. Overview of framework, contributions. -- summary of work done to tackle the challenges.  
To enhance the discriminative ability of the AutoEncoder~\cite{mcclelland1986parallel}, we propose \textbf{T}ransformation-based \textbf{E}mbedding learning of \textbf{N}ovelty \textbf{D}etection (\textbf{TEND}) to distinguish intra-class OOD inputs in an unsupervised fashion. 
% TEND is a hybrid model which combines both autoencoder-based and classifier-based approaches. 
% The overview of TEND pipeline is shown in Fig.~\ref{fig:model}, 
Based on the vanilla AutoEncoder~\cite{mcclelland1986parallel} model to learn the ``normality" of ID data in the first stage and function as a feature extractor in the second stage, TEND utilizes distorted images generated by adding transformations on the ID data, and treats the data as non-ID data (marginal OOD, see Sec.~\ref{Transformations}). A binary classifier of TEND is trained with the ID data as normal class and the  non-ID data as OOD class. Hence, the classifier is aware of the existence of outliers and gains certain identification ability of true outliers during inference without being trained on any true OOD data. To further separate OOD data from the ID ones, we learn a distance metric objective to encourage clustering of ID data during training  and enforce a margin between OOD versus ID data in the embedding space.
% (see the ``Training" part of Fig.~\ref{fig:example}) (see the ``Inference" part of Fig.~\ref{fig:example})
In summary, the main contributions of our paper are as follows:

1) We propose a new novelty detection model TEND that utilizes the AutoEncoder's feature extraction and adds discrimination ability for outliers with transformations of in-distribution data and embedding distance as auxiliary. No out-of-distribution data is required for training the model.

2) Although there have been a lot of anomaly detection research work done, the accurate detection performance results are lacking.
We compare and report the novelty detection performance details of the unsupervised TEND model with state-of-the-art anomaly detection models and one supervised model on three public medical image datasets following two experimental settings - one-vs-rest and rest-vs-one.  

3) We validate our method on diverse image datasets and demonstrate our model's effectiveness. Extensive evaluations include the detection of intra-class out-of-distribution data from the original datasets and the corresponding generated with unused transformations on in-distribution data. Given the experimental observations, our model will be beneficial in discovering new anomaly cases in medical applications without any preconceived OOD training data. 

\section{Background}\label{sec:background}

There have been a lot of research works that summarize state-of-the-art anomaly detection methods~\cite{chalapathy2019deep:chala,lee2018simple:lee,ouyang2020video:video,hendrycks2018deep:hendry,cao2020benchmark:benchmark,di2019survey:di},  generally the methods aiming for anomalous image data detection can be divided into the following three categories:  

\noindent \textbf{i. AutoEncoder-based methods}: AutoEncoder~\cite{mcclelland1986parallel} (AE) models can help extract significant embedding features by reconstructing the original images unsupervised. Trained with ID data, the architectures learn the ``normality"  and should lead to large reconstruction error when working on OOD dataset. Thus, the reconstruction error acts as the anomaly score to separate ID and OOD data~\cite{sakurada2014anomaly:autoencoder,zhou2017anomaly:autoencoder,beggel2019robust:autoencoder}. However, AutoEncoder risks learning the identity function by simply outputting the original inputs, which largely limits its discriminative ability of anomalies. Other improved versions of AE are also used for anomaly detection~\cite{tagawa2015structured,pol2019anomaly,lupo2019variational,an2015variational}, \textit{e.g.,} Variational Autoencoders (VAE)~\cite{an2015variational:an} provides probabilistic way of describing the latent space to reconstruct input data. Nevertheless, the reconstruction is often blurry and not good enough for clear discrimination of outliers. Since TEND is designed based on AutoEncoder, we take the vanilla AE~\cite{mcclelland1986parallel} as a baseline. Besides, we also compare the performance with an extension of AE that adds a Gaussian Mixture Model (GMM) head on the AutoEncoder backbone, (AE\_GMM for simplicity's sake) and the standard VAE~\cite{an2015variational:an} model. Similar with VAE, UAV-AdNet~\cite{bozcan2020uav} uses the Kullback-Leibler divergence to regularize losses for anomaly detection but focuses on autonomous surveillance systems with GPS label used, which does not apply to this work.      

\noindent \textbf{ii. Generative adversarial network (GAN) based methods}: Similar to the AE models, GAN~\cite{goodfellow2020generative} framework can also learn latent feature representations by training a fake image generator and a real-vs-fake image discriminator~\cite{zenati2018efficient:zenati,perera2019ocgan:ocgan}. With the adversarial feature learning, GAN-based anomaly detectors can acquire discriminative latent features that can be used for separating the ID data from the OOD data. To further improve the discriminative ability of latent representations, BiGAN~\cite{donahue2016adversarial:donahue} adopts a bidirectional mapping learning. GANomaly~\cite{akcay2018ganomaly:akcay} minimizes the distance of the ID data and the generated ones in latent feature space to detect the OOD data with large distance. 
% {SteGANomaly\cite{baur2020steganomaly} embeds anomaly detection into a CycleGAN-framework to perform brain tumor anomaly detection, limiting its general applications.} 
Even so, the performance of GAN-based anomaly detectors largely depends on the training of GAN models, which always require large amounts of training data for OOD and often fail to handle inputs with large image size. Instead of selecting AnoGAN~\cite{schlegl2017unsupervised:schlegl}, which detects pixel-wise anomalies rather than in image-level, we compare TEND with GANomaly~\cite{akcay2018ganomaly:akcay} and f-AnoGAN~\cite{schlegl2019f}, AnoGAN's extension, for experiments given the better performance.

\noindent \textbf{iii. Classifier-based methods}: As the novelty detection in medical images can be reduced to a one-class classification (OCC)~\cite{khan2014one} problem with the one-vs-rest setup, one-class classifiers are often used for identifying unseen classes, \textit{e.g.,} OC-SVM~\cite{scholkopf2001estimating:ocsvm}, FCDD~\cite{liznerski2020explainable:explainable}, DOC~\cite{perera2019learning:perera}, DeepSVDD~\cite{ruff2018deep:ruff}. With only ID data as training inputs, one-class classifiers often optimize a kernel-based objective function and minimize a hyper-sphere to threshold out the anomaly data based on distance. 
% With external image datasets as defined outliers, DOC~\cite{perera2019learning:perera} tries to improve the descriptiveness of the in-distribution class of the model. 
In common, the one-class classifiers exploit in-distribution data with specific object functions to threshold out anomalies. Nonetheless, their detection abilities on intra-class OOD data are not effective as the intra-class OOD data shares a lot of similarity with the ID data. Except for the one-class classifiers, ODIN~\cite{liang2017enhancing:liang} works on multi-classes datasets by adding perturbations of the input and temperature scaling to the score function to distinguish in-distribution and OOD data. Despite the efficiency and sophisticated methodology, the prerequisites of multiple OOD classes of the dataset is not typical in the medical image area and thus classifier-based methods have limited applicability in healthcare. To showcase the performance difference, we choose DeepSVDD~\cite{ruff2018deep:ruff}, which is a representative model, to compare with TEND.

\section{Method}
%Method
% overview of the entire architecture. Introduce autoencoder, classification. 

% 3.1 transformation (could be augmentation in feature space)
% 3.2 Distance metric learning
% 3.3 Joint training. Autoencoder (could be VAE) + 3.2 DML + classification loss
% 3.4 Inference (optional). Remove reconstruction. 

% comment: 
% 1. simplify annotations of the figure

\begin{figure}[tp]
\begin{center}
  \includegraphics[width=0.8\linewidth]{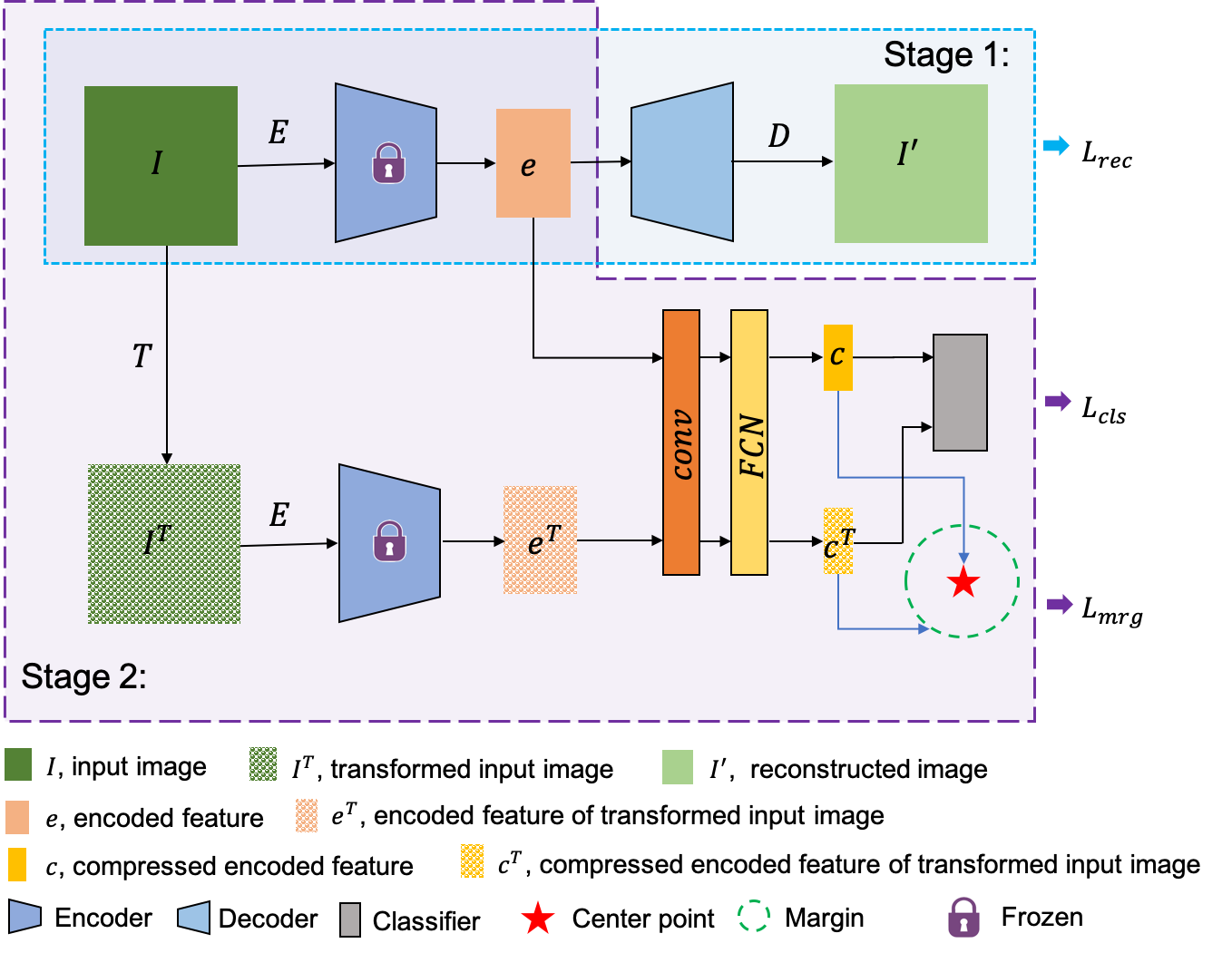}
\end{center}
  \caption{Network architecture of TEND. - Stage 1: Training AutoEncoder with in-distribution data; Stage 2: Joint training of the classifier and the margin learner.} 
\label{fig:model}
\end{figure}

TEND focuses on novelty identification for medical images. By following the one-vs-rest setup~\cite{liznerski2020explainable:explainable} and its revsered version - the rest-vs-one setup, one or more certain classes of the datasets in use are treated as normal classes. Unsupervised learning of feature embeddings for the normal classes is the fundamental step for outlier detection. GANs and AEs are all good options for this work. Nonetheless, GANs often require large amounts of data for training and are unstable for large images, we choose the vanilla AE~\cite{goodfellow2020generative} to encode the ID data. Moreover, as introduced in Sec.~\ref{sec:background}, AEs are designed for compressing inputs and have no strong discriminative ability, which makes them inappropriate for medical novelty detection because of the minute intra-class variations of medical image datasets. Thus, to enhance the discriminative ability of TEND, we train a binary classifier and a margin-aware objective function (also called margin learner) jointly to separate the normal class data from the anomalies. 

\subsection{Architecture}
Figure~\ref{fig:model} shows the network architecture of TEND, which is a two-stage novelty detector with an AutoEncoder~\cite{mcclelland1986parallel} as the feature extractor backbone. In order to train the feature extractor with only ID data, the AutoEncoder~\cite{mcclelland1986parallel} model (shown in the dotted blue box of Fig.~\ref{fig:model}) is optimized with a reconstruction loss function $L_{rec}$. The learnt bottleneck section will be frozen as indicated by the purple lock in Fig.~\ref{fig:model} and used for encoding/extracting image features in the second stage. To train the following binary discriminator without OOD data available, we add transformations on the original images to construct distribution-shifted OOD samples based on the observation that some augmentations can be useful for OOD detection by considering them as fake OOD data~\cite{tack2020csi:tack}. The details of how to construct the transformations are explained in Sec.~\ref{Transformations}. The generated OOD data should be first fed to the trained encoder to obtain the corresponding deep features. Both of the encoded features of normal and  transformed data are fed to the classifier simultaneously. With a convolutional (\textit{conv}) layer and a fully connected layer (\textit{FCN}), the classifier learns to identify the in-distribution data as normal class and the transformed images as outliers. A latent decision boundary between the two classes is optimized, the detection on true anomaly data is still not promising given the fact that the transformed images can not represent the true outliers' distribution. The decision boundary may not work for the anomalies in the feature space. To solve this problem, TEND adopts the margin-aware learning idea of DeepSVDD~\cite{ruff2018deep:ruff} to optimize a distance objective function simultaneously. Different from the objectives only for ID data~\cite{ruff2018deep:ruff}, TEND works on both the ID data and the fake OOD data by enforcing the embeddings of ID data to cluster around a voted center $O$ (see Sec.~\ref{jointraining} for more details) whereas pushing out the embeddings of the generated abnormal class with a predefined margin $R$.   
% and subsequently go to a two-layer perceptron with ReLU non-linearity. 

\subsection{Transformations for generating fake OOD data}\label{Transformations}
\begin{figure*}
\begin{center}
    \includegraphics[width=\linewidth]{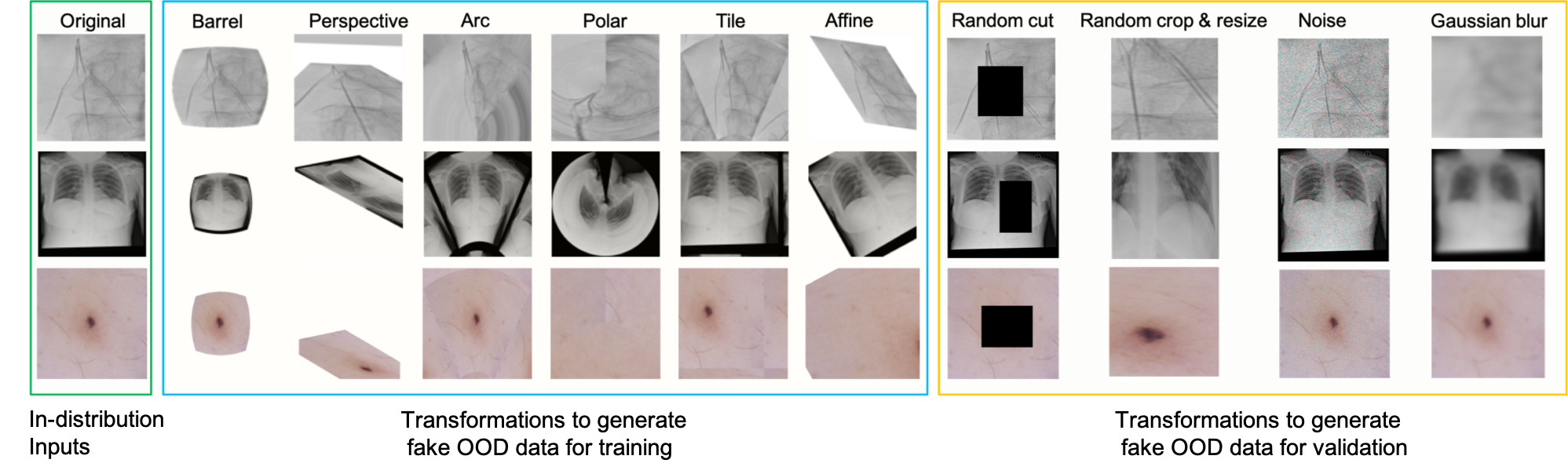}
\end{center}
  \caption{Examples of transformations used for generating fake OOD data. Three image examples from IVC-Filter (1st row), RSNA (2nd row) and ISIC2019 (3rd row) datasets are presented. The original data in the green box are inputs from in-distribution class, the transformed in-distribution images in the blue box are auxiliary data as anomalies feed to TEND's classifier during training, other possible transformations shown in the yellow box are for validation. } 
\label{fig:example}
\end{figure*}
SimCLR~\cite{chen2020simple} has performed an extensive study on which family of augmentations leads to a better self-supervised learning, i.e., which transformations should be considered as positives. The authors report that some of the examined augmentations (e.g., rotation), could lead to degraded performance. Based on the observation, such augmentations can be useful for OOD detection by considering them as fake OOD data. Therefore, we leverage a family of transformations
% SimCLR~\cite{chen2020simple} proposed to generate augmentations with transformations and learn representations by maximizing agreement between differently augmented views of the same data example via a contrastive loss in the latent space. This helps increase object recognition ability for more general natural image tasks, which is not suitable for medical OOD detection as the detector should be sensitive to suspicious inputs, including noisy and cropped images and no OOD data can be used for model training. 
% Note that the augmentations used in SimCLR are random cropping, random color distortions, and random Gaussian blur, which will not change the object visual appearance. Inspired by the augmentation usage but not serving for the same purpose in our work, 
and utilize more complex transformations and distortion functions that will change the visual features of the original inputs to generate fake abnormal data for training in OOD model. The generated auxiliary data are fed to the forehead of the TEND backbone and then to the classifier, which helps separate the embedding features of the ID data from those of the unknown OOD data. Different from the most common transformations, \textit{e.g.,} rotation, used in classic data augmentation, we adopt a range of different distortions, \textit{i.e., barrel, perspective, arc, polar, tile, affine} defined in the \textit{Image.distort} method of \textit{Wand} package\footnote{https://docs.wand-py.org/en/0.6.5/guide/distortion.html}. The blue box in the middle part of Fig.~\ref{fig:example} shows the six different transformations on the three datasets. These transformations bring significant difference to the original inputs and generates intra-class OOD samples. We treated these extreme distortions of ID data as outliers for training. Expect for the six distortions used in this paper, there are more transformations worthwhile being explored. To further demonstrate the benefits of training the TEND model using extreme transformations, we use moderate distortions, such as randomly cutting, randomly cropping and resizing, addition of noises, Gaussian blurring only for validation (shown in the right yellow box of Fig.~\ref{fig:example}). The package usage and parameters selection for the six training distortions and the four validation transformation are present in our code repository.

\subsection{Joint training}~\label{jointraining}
With an AutoEncoder~\cite{mcclelland1986parallel} as the backbone, TEND incorporates a classifier and a margin-aware embedding mapping to gain discriminative ability for anomalies. 
In the first stage, the backbone is trained only on ID data. Suppose that the input image $I$ and reconstructed image $I^{'}$ is with size of $M\times N$, a reconstruction objective $f_{rec}$ defined in Eqn.~\ref{rec} is used to optimize the learning embedding representations of the normal class. This first-stage training ensures the feature extractor to focus on learning the ``normality" of in-class data. 

\begin{equation}\label{rec}
f_{rec}=min\frac{1}{M}\frac{1}{N}\sum_{i=1,j=1}^{M,N}\left \|I_{ij}-I^{'}_{ij}  \right \|^{2}
\end{equation}

\begin{equation}\label{cls}
    L_{cls} = \frac{1}{S} \sum_{i=1}^{S} {y_{i}}\cdot log(p(y_{i})) + (1-y_{i})\cdot log(1-p(y_{i}))
\end{equation}
With the distorted ID data as anomalies in the second stage, the binary discriminator is able to train with a final output indicating the data class. Notably, the inputs of this classifier are the encoded features extracted by the backbone. Here, the AutoEncoder model is fully frozen and only used for extracting image features. The encoded features $e, e^{T}$ are processed by a following convolutional layer (conv) and a fully connected layer (FCN) of the classifier. Thus, the embeddings learnt by the encoder are mapped to a new compressed space as $c, c^{T}$ with size of $K$ (512 in our case). The classifier enables the separation of the compressed features of the ID data and the distorted data. A binary cross entropy loss function $L_{cls}$ shown in Eqn.~\ref{cls} is utilized for optimizing, with the $S$ to be the total number of the training data, $y_{i}$ representing the $ith$ data's binary label and $p(y_{i})$ being the corresponding probability of the prediction. Nonetheless, the transformations $T$ can only introduce limited class variations, hence the identification for real OOD data is still not ideal. Thus, a margin-aware objective is jointly trained to force the clustering of the compressed features of the ID data and the surrounding of the transformed ID data outside the margin as illustrated by Fig.~\ref{fig:example}. 

In experiments, we test three margin $R$ values (150, 250 and 500). Similar to DeepSVDD~\cite{ruff2018deep:ruff}, the compressed feature center \textit{O} is calculated by the mean of all the ID data's compressed features. Before calculation, TEND's classifier block is trained with several warm-up epochs, (\textit{e.g.,} 10 epochs), then the center \textit{O} is defined with the same size of \textit{K} as the compressed feature \textit{c}. Since then, the margin learner of TEND is trained together with the discriminator. Importantly, the margin learner has different learning objectives for the normal class ($g_{in}$) shown in Eqn.~\ref{mrgiid} and the generated abnormal class ($g_{out}$) shown in Eqn.~\ref{mrgtfm}. 
\begin{equation}\label{mrgiid}
g_{in}=min\frac{1}{K}\sum_{i=1}^{K}\left \|c_{i}-O  \right \|^{2}
\end{equation}
\begin{equation}\label{mrgtfm}
g_{out}=min\frac{1}{K}\sum_{i=1}^{K}max(R-\left \|c^{T}_{i}-O  \right \|^{2}, 0)
\end{equation}

In summary, TEND has two stage-wise losses. The first-stage loss is for the reconstruction of the AutoEncoder training, \textit{i.e.,} $L_{1st}=L_{rec}$. The second-stage loss includes the binary classifier and the margin learner, \textit{i.e.,} $L_{2nd} = L_{cls}+L_{mrg}$. In experiments, we use mean square error (MSE) loss for $L_{rec}$ and binary cross entropy (BCE) loss for $L_{cls}$. Marginal loss $L_{mrg}$ equals the summation of the mean of distance errors for ID data and the mean of the errors for distorted data.

\subsection{Implementation details}
An AutoEncoder architecture is trained as our baseline, the trained model later on is treated as the backbone of TEND. 
We report the encoder, decoder, Conv, FCN parts of TEND in Table~\ref{tab:ae}.
\textit{FC} is fully connected layer, \textit{Conv} stands for the convolutional layer , \textit{TConv} means the transposed convolutional layer. \textit{channel} indicates the image channel. All the \textit{Conv} and \textit{TConv} layers use kernel filter size 4, stride 2 and padding 1. The encoder encodes input images as \textit{e}, while the \textit{Conv} layer compresses \textit{e} to \textit{c} with smaller sizes. Each \textit{Conv} and \textit{TConv} is followed by a standard batch-normalization layer and a relu function.

\begin{table*}[!ht]
\centering
\caption{TEND architecture details.}
\label{tab:ae}
\begin{tabular}{|l|c|c|c|c|} 
\hline
\textbf{Dataset}  & \textbf{Encoder}   & \textbf{Decoder}    & \textbf{Conv}    & \textbf{FCN}   \\ \hline
\textit{\begin{tabular}[c]{@{}l@{}}IVC-Filter/\\RSNA/\\ISIC\end{tabular}} & \textit{\begin{tabular}[c]{@{}c@{}}Conv(\textit{channel},16)\\Conv(16,32)\\Conv(32,64)\\Conv(64,128)\\Conv(128,256)\end{tabular}} & \textit{\begin{tabular}[c]{@{}c@{}}TConv(256,128)\\TConv(128,64)\\TConv(64,32)\\TConv(32,16)\\TConv(16,\textit{channel})\end{tabular}} & \textit{Conv(256,512)} & \textit{\begin{tabular}[c]{@{}c@{}}FC(2048,512)\\FC(512,1)\end{tabular}}  \\
\hline
\end{tabular}
\end{table*}

In our experiments, we use Adam optimizer with a learning rate of 0.001 for model training. Each network is trained with 50-150 epochs depending on the dataset size and the data complexity as datasets with more complex data or large amounts of samples often take more time to get the loss decreased to a satisfactory level. When training with the margin-aware metric, we run 10 warm-up epochs first and then calculate the embedding center \textit{O}. The pipelines are developed using Pytorch 1.5.0, Python 3.0. and Cuda compilation tools V10.0.130 on a machine with 3 NVIDIA Quadro RTX 6000 with 24GB memory.

\subsection{Anomaly score}
As a standard evaluation procedure for anomaly detectors, the ID and outliers are mixed for computing the accuracy while different detectors have different anomaly score definitions. For the baseline AutoEncoder model, we set the reconstruction error as the OOD data score. TEND does not focus on the reconstruction, therefore, the final anomaly score of TEND is the classification probability adding the marginal distance. Giving the fact that the classification probability $p$ is in range $[0-1]$ while the distance value $d$ is in $[0,+\infty)$, we scale down the distance value $d$ by dividing the predefined margin $R$, \textit{i.e.,} $d^{'} ={\frac{d}{R}}$. Therefore, the final anomaly score for TEND is $S_{i}=\lambda p_{i}+(1-\lambda)d^{'}_{i}$. The value of $\lambda$ is set as 0.5 in our experiments as default. To further demonstrate the effectiveness of each component of TEND, we have done the ablation study of TEND and reported the results in Sec.~\ref{ablation}. TEND without the binary classifier is called \textit{MarginLearner} (the anomaly score is $d^{'}$). 
% TEND without the margin learner is called \textit{TEND\textbf{w/o}mrg} (the anomaly score is \textit{p}), and 

%showcase
\subsection{Evaluation metrics}~\label{eval_metrics}
Having the anomaly prediction score, the detection accuracy largely depends on the threshold setting. To be fair, the detection evaluation should be threshold-invariant. Following the standard evaluation metrics used in other works~\cite{yu2019unsupervised,liang2017enhancing}, we adopt AUROC (AUC in short) to showcase the performance difference among the models. AUROC is the Area Under the Receiver Operating Characteristic curve, which is a threshold independent metric. The AUROC can be interpreted as the probability that a positive example is assigned a higher detection score than a negative example. To find an optimal threshold for receiver operating characteristic (ROC) curve by tuning the decision thresholds, we use the Geometric Mean (G-Mean) as the metric to determine the best threshold values and report the resulted true positive rate ($TPR = {\frac{TP}{TP+FN}}$) and false positive rate ($FPR = {\frac{FP}{FP+TN}}$). The difference between the TPR and FPR given the optimal selection, $DIFF=TPR-FPR$, is also reported for model comparison. Large differences stand for better true and false positive predictions. Moreover, we measure the uncertainty of models' performance with 10 rounds of bootstrapping estimations, by randomly sampling the predictions to the same amount of test samples with replacement. The resulting standard deviation values are present in \cref{tab:acc,tab:acc000,tab:val,tab:val0000}.

\section{Experiments}
% + quantitative results / + qualitative results / + ablation studies.
In this section, we perform empirical evaluations of TEND on publicly available medical image datasets with varying complexity. For evaluating the accuracy in identifying novel class data, we compare our results with state-of-the-art unsupervised OOD models, starting from simple vanilla AutoEncoder (AE)~\cite{mcclelland1986parallel} model and a variational AutoEncoder (VAE)~\cite{an2015variational:an}, to DeepSVDD~\cite{ruff2018deep:ruff}, GANomaly~\cite{akcay2018ganomaly:akcay}, f-AnoGAN~\cite{schlegl2019f} models. We also compare our unsupervised TEND model against a supervised binary classifier which was trained on both ID and OOD data for the detection task. 

\subsection{Datasets}
In our experiments, we have three medical datasets in use, including inferior vena cava (IVC) filters on radiographs~\cite{ni2020deep:ivc} and RSNA chest x-ray dataset~\cite{wang2017chestx}, ISIC2019~\cite{codella2018skin}. IVC-filter dataset has 14 classes in total. The details are ALN (73 images), BardSimonNitinol (59 images), Optease (129 images), BardDenali (50 images), Celect (75 images), Option (196 images), BardEclipseG2X (84 images), CelectPlatinum (48 images), Trapease (100 images), BardG2 (45 images), Greenfield12Fr (122 images), Tulip (99 images), BardMeridian (55 images), GreenfieldTitanium (101 images). RSNA has 3 classes - normal, with opacity, not normal in total. ISIC2019~\cite{codella2018skin} consists of 8 classes, \textit{i.e.,} Melanoma (MEL, 4148 images), Melanocytic nevus (NV, 11559 images), Basal cell carcinoma(BCC, 3323 images), Actinic keratosis (AK, 867 images), Benign keratosis (BKL, 2240 images), Dermatofibroma (DF, 239 images), Vascular lesion (VASC, 253 images), Squamous cell carcinoma (SCC, 628 images). The IVC-filter and ISIC2019 image are with varying sizes, with the width size ranging from 150 to 1500, height size ranging from 150 to 1500 roundly, \textit{e.g.,} $469\times365\times3$. The RSNA dataset is in dicom format, each dicom file has the pixel array of size $1024\times1024$. To unify the training pipeline, we resize all the IVC-Filter, RSNA and ISIC data in $256\times256\times channel$. 

For the one-vs-rest setting, the in-class and rest classes data details are summarized in Table~\ref{tab:dataset}. Due to the data imbalance, we usually pick the class with the most data as our in-distribution data and all the left classes as intra-class OOD data. For IVC-filter, we select the \textit{Option} type as the normal class; for RSNA dataset, we treat the \textit{normal} class as ID data; for ISIC2019 dataset, we choose the NV class with the most samples as ID inputs. The total numbers of ID and OOD data for each dataset are reported in the column of \textbf{\#images} in Table~\ref{tab:dataset}. Notably, the rest-vs-one setting experiments treat the classes conversely.

\begin{table*}[tp]
\centering
\caption{Three publicly available dataset used in the study - total number of images in the dataset, In-distribution data ($D_{in}$) and out-of-distribution data ($D_{out}$) with one-vs-rest setting.}
\label{tab:dataset}
\begin{tabular}{|l|c|c|c|c|c|} 
\hline
\multirow{2}{*}{\textbf{Dataset}} & \multirow{2}{*}{\textbf{total classes}} & \multicolumn{2}{c|}{\textbf{$D_{in}$}} & \multicolumn{2}{c|}{\textbf{$D_{out}$}}   \\ 
\cline{3-6}
                         &                              & \textbf{class}  & \textbf{\#images}        & \textbf{class}        & \textbf{\#images}  \\
\hline
IVC-Filter~\cite{ni2020deep:ivc}               & 14                           & \textit{Option}     &   196             & \textit{\begin{tabular}[c]{@{}l@{}}BardSimonNitinol, ALN...\end{tabular}}     &       1,040      \\ 
\hline
RSNA~\cite{wang2017chestx}                     & 3                            & \textit{normal} &     8,851             &  \textit{\begin{tabular}[c]{@{}l@{}}with opacity, not normal\end{tabular}}   & 21,376  \\ 
\hline
ISIC~\cite{codella2018skin}                 & 8                          & \textit{NV}  &   11,559         & \textit{MEL, BCC...} &  11,698  \\ 
\hline
\end{tabular}
\end{table*}

\subsection{Training and evaluation settings}\label{train_eval}
To train and evaluate OOD detectors' performance, we split the in-distribution data with 80\% as training set $D_{in}^{tr}$ and 20\% as test set $D_{in}^{te}$ and use all the left classes as $D_{out}$. For OOD detection evaluation, we mixed $D_{in}^{te}$ and $D_{out}$ by assigning the ID data with label 0 and OOD data with label 1. Since this paper focuses on intra-class OOD detection, we will report the OOD detection results within the same dataset instead of crossing different datasets.

\begin{table*}[tp]
\caption{FPR, TPR values, difference of TPR and FPR values, and AUC scores of various OOD detection methods trained on IVC-Filter~\cite{ni2020deep:ivc}, RSNA~\cite{wang2017chestx} and ISIC2019~\cite{codella2018skin} datasets \textbf{with the one-vs-rest setting}. Bold numbers are the best results and underlined numbers are the second best. Models with * are supervised and those without * are unsupervised.}
\label{tab:acc}
\centering
\resizebox{\textwidth}{!}{%
\begin{tabular}{|l|cccc|cccc|cccc|cccc|}
\hline
\multirow{2}{*}{Methods}   & \multicolumn{4}{c|}{IVC-filter} &\multicolumn{4}{c|}{RSNA} &\multicolumn{4}{c|}{ISIC2019}  \\ \cline{2-13} 
& $\downarrow$\textit{FPR}     & $\uparrow$\textit{TPR}  & $\uparrow$\textit{DIFF}   & $\uparrow$\textit{AUC}    & $\downarrow$\textit{FPR}       & $\uparrow$\textit{TPR}    & $\uparrow$\textit{DIFF}  & $\uparrow$\textit{AUC}    & $\downarrow$\textit{FPR}       & $\uparrow$\textit{TPR}   & $\uparrow$\textit{DIFF}   & $\uparrow$\textit{AUC}   \\ \hline 
AutoEncoder~\cite{mcclelland1986parallel}      & {$0.198\pm0.104$}     & {$0.350\pm0.075$} & {$0.152\pm0.067$}   & {$0.436\pm0.040$}  & {$0.318\pm0.014$}   & {$0.461\pm0.009$} & {$0.143\pm0.010$} & {$0.566\pm0.004$}    &{$0.833\pm0.060$}  & {$0.186\pm0.059$}  & {$-0.648\pm0.025$}  &{$0.096\pm0.003$} \\
AE\_GMM     & {$0.224\pm0.138$}    & {$0.153\pm0.008$} & {$-0.071\pm0.134$}   & {$0.464\pm0.067$}  & {$0.496\pm0.012$}  & {$0.321\pm0.003$} & {$-0.175\pm0.013$} & {$0.412\pm0.006$}  & {$0.083\pm0.006$} & {$0.211\pm0.003$} & {$0.128\pm0.006$} & {$0.564\pm0.003$} \\
VAE~\cite{an2015variational:an} & {$0.489\pm0.097$}      & {$0.707\pm0.076$}   & {$0.218\pm0.117$}  & {$0.542\pm0.080$}   & {$0.473\pm0.001$}   & {$0.462\pm0.001$} & {$-0.011\pm0.012$} & {$0.487\pm0.001$}    & {$0.351\pm0.011$} & {$0.395\pm0.007$} & {$0.045\pm0.007$} & {$0.471\pm0.005$} \\ 
MarginLearner & {$0.426\pm0.099$}     & {$0.549\pm0.033$}  & {$0.123\pm0.098$}  & {$0.568\pm0.055$}  & {$0.475\pm0.016$}   & {$0.478\pm0.013$} & {$0.003\pm0.010$} & {$0.491\pm0.005$}    & {$0.517\pm0.020$} & {$0.584\pm0.024$} & {$0.067\pm0.010$} & {$0.530\pm0.005$} \\
DeepSVDD~\cite{ruff2018deep:ruff}        & {$0.503\pm0.106$}     & {$0.672\pm0.042$} & {$0.170\pm0.130$}   & {$0.500\pm0.075$}  & {$0.508\pm0.021$}      & {$0.413\pm0.023$}  & {$-0.095\pm0.015$} & {$0.421\pm0.009$}  &{$0.348\pm0.021$}& {$0.621\pm0.021$} & {\underline{0.273$\pm$0.006}} & {$0.677\pm0.003$} \\
GANomaly~\cite{akcay2018ganomaly:akcay}       & {$0.446\pm0.172$}     & {$0.627\pm0.227$} & {$0.181\pm0.200$}   & {$0.518\pm0.103$}  & {$0.524\pm0.005$}   & {$0.678\pm0.015$} & {$0.154\pm0.009$}  & {$0.576\pm0.005$} & {$0.396\pm0.030$} & {$0.481\pm0.027$} & {$0.086\pm0.012$} & {$0.551\pm0.009$}\\
f-AnoGAN~\cite{schlegl2019f} & {$0.419\pm0.077$}     & {$0.511\pm0.070$} & {$0.092\pm0.045$}   & {$0.544\pm0.022$} & {$0.365\pm0.033$}  & {$0.541\pm0.029$}  &  {\underline{0.176$\pm$0.008}}  & {$0.614\pm0.005$}  &  {$0.366\pm0.007$} & {$0.600\pm0.007$} & {$0.234\pm0.005$}  & {$0.647\pm0.003$} \\
\hline
TEND\_150 (ours)    & {$0.219\pm0.077$}        & {$0.749\pm0.086$}  & {\textbf{0.531$\pm$0.071}}     & {\textbf{0.772$\pm$0.030}}  & {$0.425\pm0.029$}      & {$0.590\pm0.026$}  & {$0.165\pm0.010$}    & {\underline{0.615$\pm$0.006}}      & {$0.377\pm0.016$} & {$0.596\pm0.015$} & {$0.220\pm0.009$} & {$0.650\pm0.006$} \\
TEND\_250 (ours)    & {$0.160\pm0.091$}        & {$0.684\pm0.035$}  & {\underline{0.524$\pm$0.082}}     & {$0.752\pm0.051$}  & {$0.389\pm0.045$}       & {$0.561\pm0.043$}   & {$0.172\pm0.009$}  & {\underline{0.615$\pm$0.006}}       & {$0.326\pm0.017$} & {$0.669\pm0.020$} & {\textbf{0.343$\pm$0.011}} & {\textbf{0.717$\pm$0.006}} \\ 
TEND\_500 (ours)    & {$0.122\pm0.099$}        & {$0.639\pm0.095$}  & {0.517$\pm$0.042}     & {\underline{0.760$\pm$0.028}}  & {$0.438\pm0.040$}       & {$0.616\pm0.041$}   & {\textbf{0.178$\pm$0.008}} & {\textbf{0.627$\pm$0.005}}       & {$0.351\pm0.012$} & {$0.618\pm0.011$} & {$0.268\pm0.009$} & {\underline{0.678$\pm$0.006}} \\ \hline
BinaryClassifier* & {$0.280\pm0.006$}    & {$0.847\pm0.003$}  & {\textbf{0.567$\pm$0.006}}  & {\textbf{0.853$\pm$0.003}}  & {$0.417\pm0.007$}   &{$0.589\pm0.006$}  & {$0.172\pm0.008$} & {$0.593\pm0.003$}    & {$0.497\pm0.023$} & {$0.340\pm0.015$} & {$-0.157\pm0.010$} & {$0.363\pm0.004$} \\ \hline
\end{tabular}%
}
\end{table*}

\subsection{Quantitative results}
\subsubsection{One-vs-rest results}
Following the one-vs-rest setting, Table~\ref{tab:acc} presents the AUC scores and the corresponding FPR, TPR values determined by the optimal thresholds for AutoEncoder~\cite{mcclelland1986parallel}, VAE~\cite{an2015variational:an}, DeepSVDD~\cite{ruff2018deep:ruff}, GANomaly\cite{akcay2018ganomaly:akcay}, f-AnoGAN~\cite{schlegl2019f} and TEND models with margin 150 (\textit{i.e.}, TEND\_150), 250 (\textit{i.e.}, TEND\_250) and 500 (\textit{i.e.}, TEND\_500). The difference between the TPR and FPR is also reported in the \textit{DIFF} column in Table~\ref{tab:acc}. $\downarrow$ means the lower the value the better the model is while $\uparrow$ stands for the higher the value the better the model performs. Thus, we expect the model to have high AUC score and prefer low FPR and high TPR values when deploying the models with the optimal threshold as decision boundary, which means the larger the difference between TPR and FPR the better. The best and 
second best \textit{DIFF} and AUC results are highlighted by bold and underline respectively. {Among the unsupervised anomaly detectors, our model TEND\_150 attains the optimal \textit{DIFF} result 0.531  and AUC score 0.772 for IVC-Filter dataset and second best AUC score 0.615 for RSNA dataset; TEND\_250 achieves the second highest \textit{DIFF} 0.524 for IVC-Filter dataset and the second highest AUC score 0.615 for RSNA dataset. Meanwhile, TEND\_250 reaches the best \textit{DIFF} 0.343 and AUC score 0.717 for ISIC2019 dataset compared to other methods. TEND\_500 reaches the sub-optimal AUC score 0.760 for IVC-Filter dataset; has the largest \textit{DIFF} value 0.178 and AUC score 0.627 for RSNA dataset; obtains the second best AUC score 0.678 for ISIC2019. GANomaly performs better than DeepSVDD on IVC-Filter and RSNA datasets with higher \textit{DIFF} and AUC values, while DeepSVDD exceeds GANomaly on ISIC2019 dataset. Across the three datasets, f-AnoGAN generally outperforms GANomaly and its performance gradually improves as the training dataset becomes larger. Nevertheless, our model TENDs show certain advantages in acquiring better accuracy and exhibits competitive performances compared with other unsupervised models. Notably, we implement TEND with three different margins to show the difference with changing settings. By observing our results in Table~\ref{tab:acc}, no unique margin in TEND provides the optimal result on all the datasets and thus it needs to be tuned for specific experiments. The effects of applying different radiuses are present in Sec.~\ref{qual}. The MarginLearner and the supervised model BinaryClassifier are also discussed in ablation study (see Sec.~\ref{ablation}).}

\subsubsection{Rest-vs-one results}
To further compare the models' performances, the complementary experimental setting - rest-vs-one is implemented with the results reported in Table~\ref{tab:acc000}. Same as the one-vs-rest experiments, we keep the tested models consistent, and change the in-distribution class as OOD classes and the previous OOD data as our in-distribution data. The training and testing processes are the same as reported in Sec.~\ref{train_eval}. {Our model TEND\_150 gets the best \textit{DIFF} 0.291 and AUC score 0.650 for IVC-Filter dataset, and obtains the sub-optimal \textit{DIFF} 0.126 and
AUC score 0.584 for RSNA dataset.} GANomaly performs the best for RSNA dataset. TEND\_250 reaches the sub-optimal results for ISIC2019 dataset whereas f-AnoGAN can achieve the best.Generally the detection of anomalies under rest-vs-one setting is more challenging than the one-vs-rest setting and nearly no model can work well for all the situations. Still, TEND has satisfactory performances across the three datasets with the rest-vs-one setting. 

\begin{table*}[!tp]
\caption{FPR, TPR values, difference of TPR and FPR values, and AUC scores of various OOD detection methods trained on IVC-Filter~\cite{ni2020deep:ivc}, RSNA~\cite{wang2017chestx} and ISIC2019~\cite{codella2018skin} datasets \textbf{with the rest-vs-one setting}. Bold numbers are the best results and underlined numbers are the second best. Models with * are supervised and those without * are unsupervised.}
\label{tab:acc000}
\centering
\resizebox{\textwidth}{!}{%
\begin{tabular}{|l|cccc|cccc|cccc|cccc|}
\hline
\multirow{2}{*}{Methods}   & \multicolumn{4}{c|}{IVC-filter} &\multicolumn{4}{c|}{RSNA} &\multicolumn{4}{c|}{ISIC2019}  \\ \cline{2-13} 
& $\downarrow$\textit{FPR}     & $\uparrow$\textit{TPR}  & $\uparrow$\textit{DIFF}   & $\uparrow$\textit{AUC}    & $\downarrow$\textit{FPR}       & $\uparrow$\textit{TPR}    & $\uparrow$\textit{DIFF}  & $\uparrow$\textit{AUC}    & $\downarrow$\textit{FPR}       & $\uparrow$\textit{TPR}   & $\uparrow$\textit{DIFF}   & $\uparrow$\textit{AUC}   \\ \hline 
AutoEncoder~\cite{mcclelland1986parallel}     & {$0.706\pm0.163$}    & {$0.312\pm0.159$} & {$-0.394\pm0.059$}   & {$0.165\pm0.027$}  & {$0.760\pm0.022$}   & {$0.544\pm0.046$} & {$-0.216\pm0.024$} & {$0.321\pm0.005$}    & {$0.593\pm0.023$} & {$0.383\pm0.024$} & {$-0.210\pm0.012$} & {$0.353\pm0.007$} \\
AE\_GMM & {$0.728\pm0.053$}     & {$0.748\pm0.022$}   & {$0.020\pm0.058$}   & {$0.510\pm0.029$}   & {$0.600\pm0.008$}   & {$0.584\pm0.004$} & {$-0.016\pm0.009$} & {$0.492\pm0.005$}    & {$0.159\pm0.005$} & {$0.059\pm0.002$} & {$-0.100\pm0.006$} & {$0.450\pm0.003$} \\
VAE~\cite{an2015variational:an} & {$0.359\pm0.082$}     & {$0.464\pm0.088$}  & {$0.105\pm0.063$}  & {$0.560\pm0.035$}  & {$0.518\pm0.029$}   & {$0.453\pm0.027$} & {$-0.065\pm0.007$} & {$0.461\pm0.036$}   & {$0.518\pm0.032$} & {$0.658\pm0.045$} & {$0.140\pm0.016$} & {$0.575\pm0.005$} \\ 
MarginLearner & {$0.617\pm0.022$}    & {$0.619\pm0.045$}  & {$0.003\pm0.043$}  & {$0.484\pm0.025$}  & {$0.510\pm0.018$}  & {$0.527\pm0.016$} & {$0.017\pm0.006$} & {$0.514\pm0.004$}   & {$0.510\pm0.018$} & {$0.527\pm0.016$} & {$0.017\pm0.006$} & {$0.514\pm0.004$} \\
DeepSVDD~\cite{ruff2018deep:ruff}       & {$0.514\pm0.043$}      & {$0.475\pm0.045$}  & {$-0.039\pm0.065$}   & {$0.439\pm0.043$}   & {$0.514\pm0.028$}       & {$0.552\pm0.032$}  & {$0.038\pm0.007$} & {$0.522\pm0.004$} & {$0.530\pm0.007$} & {$0.540\pm0.013$} & {$0.010\pm0.011$} & {$0.487\pm0.007$}\\
GANomaly~\cite{akcay2018ganomaly:akcay}         & {$0.595\pm0.060$}     & {$0.622\pm0.040$} & {$0.027\pm0.051$}   & {$0.449\pm0.036$} & {$0.396\pm0.014$}   & {$0.638\pm0.014$} & {\textbf{0.242$\pm$0.004}}  & {\textbf{0.656$\pm$0.003}} & {$0.462\pm0.0166$} & {$0.583\pm0.019$} & {$0.121\pm0.009$} & {$0.570\pm0.005$}\\
f-AnoGAN~\cite{schlegl2019f} & {$0.419\pm0.077$}   &{$0.511\pm0.070$}  & {$0.092\pm0.045$}    & {$0.544\pm0.022$} & {$0.295\pm0.029$}  & {$0.276\pm0.012$}  &  {$-0.019\pm0.019$}  & {$0.406\pm0.005$}  & {$0.276\pm0.004$} & {$0.677\pm0.006$} & {\textbf{0.401$\pm$0.007}}  & {\textbf{0.718$\pm$0.004}} \\
\hline
TEND\_150 (ours)    & {$0.359\pm0.057$}        & {$0.640\pm0.031$}  & {\textbf{0.291$\pm$0.051}}     & {\textbf{0.650$\pm$0.028}}  & {$0.452\pm0.022$}       & {$0.578\pm0.024$} & {\underline{0.126$\pm$0.007}}    & {\underline{0.584$\pm$0.003}}     & {$0.336\pm0.015$} & {$0.501\pm0.006$} & {$0.164\pm0.014$} & {$0.608\pm0.007$}\\
TEND\_250 (ours)    & {$0.427\pm0.061$}         & {$0.582\pm0.071$}     & {$0.155\pm0.058$}      & {\underline{0.573$\pm$0.039}}    & {$0.492\pm0.016$}        & {$0.577\pm0.015$}     & {$0.084\pm0.006$}     & {$0.549\pm0.004$}    & {$0.386\pm0.014$}  & {$0.623\pm0.011$}  & {\underline{0.237$\pm$0.011}}  & {\underline{0.637$\pm$0.008}}  \\ 
TEND\_500 (ours)    & {$0.428\pm0.069$}      & {$0.584\pm0.081$}   & {\underline{0.156$\pm$0.038}}      & {\underline{0.573$\pm$0.025}}   & {$0.487\pm0.015$}     &{$0.550\pm0.014$}   & {$0.063\pm0.008$} & {$0.541\pm0.005$}      & {$0.412\pm0.018$} & {$0.533\pm0.016$} & {$0.121\pm0.013$} & {$0.582\pm0.009$} \\ \hline
BinaryClassifier* & {$0.617\pm0.022$}        & {$0.619\pm0.045$}   & {$0.003\pm0.043$}     & {$0.484\pm0.025$}  & {$0.510\pm0.018$}      & {$0.527\pm0.016$}    & {$0.017\pm0.006$}   & {$0.514\pm0.004$}       & {$0.471\pm0.014$} & {$0.599\pm0.017$} & {$0.128\pm0.005$} & {$0.584\pm0.004$} \\  \hline
\end{tabular}%
}
\end{table*}

\subsection{Ablation studies}\label{ablation}
To further explore the effectiveness of each module in TEND, we perform the ablation studies with the settings of removing the binary classifier from TEND (MarginLearner) and training a supervised binary classifier (BinaryClassifier) respectively. For the one-vs-rest setting, the results are shown as \textit{MarginLearner} with radius setting 150 in Table~\ref{tab:acc}, with slight \textit{DIFF} and AUC improvements compared to the baseline AutoEncoder on IVC-Filter and ISIC2019 datasets. {Comparatively, TEND\_150 enlarges the \textit{DIFF} with 0.379, 0.022 and 0.868 improvements, and increases the AUC scores by 0.336, 0.049, 0.554 respectively on IVC-Filter, RSNA and ISIC2019 datasets. For the rest-vs-one setting, compared with the \textit{MarginLearner}, TEND\_150 achieves the \textit{DIFF} with 0.288, 0.109, 0.147 improvements for IVC-Filter, RSNA and ISIC2019 dataset respectively; and enhances the AUC score with 0.166, 0.070, 0.094 for the three datasets. These observations indicate the effectiveness of TEND's architecture.}

We also report the performance of an AE extension, AE\_GMM, which clusters the embeddings from the AutoEncoder backbone and predicts the data classes - ID or OOD. From both Table~\ref{tab:acc} and Table~\ref{tab:acc000}, a GMM head can improve the discriminative ability of AutoEncoder to certain extent, however, when testing on transformed OOD data in Table~\ref{tab:val} and Table~\ref{tab:val0000}, the advantages fail to remain. In comparison, TEND's heads on AE have more generalization ability and demonstrate consistent detection performance.

Instead of training the binary classifier of TEND model in an unsupervised fashion, we include partial true OOD data in training data. Since IVC-Filter and ISIC2019 datasets have multiple classes, we randomly select 2-3 OOD classes for training and the left classes for validation. 

One-vs-rest setting:
For RSNA datasets, we use the class \textit{not normal} (see Table~\ref{tab:dataset} for details) for known OOD data and test the model on the left \textit{with opacity} data. The supervised \textit{BinaryClassifer} is also evaluated with quantitative results appended in the end of Table~\ref{tab:acc}. {With prior knowledge about OOD data, the BinaryClassifer can achieve very high AUC scores for IVC-Filter (+0.081 compared to the best of unsupervised results). Nonetheless, this advantage fails to remain on other datasets, which indicates the benefits from prior knowledge are limited.} 

Rest-vs-one setting:
For RSNA datasets, we use the class \textit{normal} as known OOD data and \textit{not normal} as ID data, the left class is used for evaluation. Different from the observation above, the corresponding results in Table~\ref{tab:acc000} for BinaryClassifier fail to exceed the unsupervised models, more results can be observed in Table~\ref{tab:val0000}. In conclusion, the supervised BinaryClassifier may lack generalization ability when dealing with unexpected data.
Please refer Sec.~\ref{nonlinear_trans} for more experimental results and discussions.

\subsection{Qualitative results}~\label{qual}
\begin{figure}[htp]
\begin{center}
  \includegraphics[width=\linewidth]{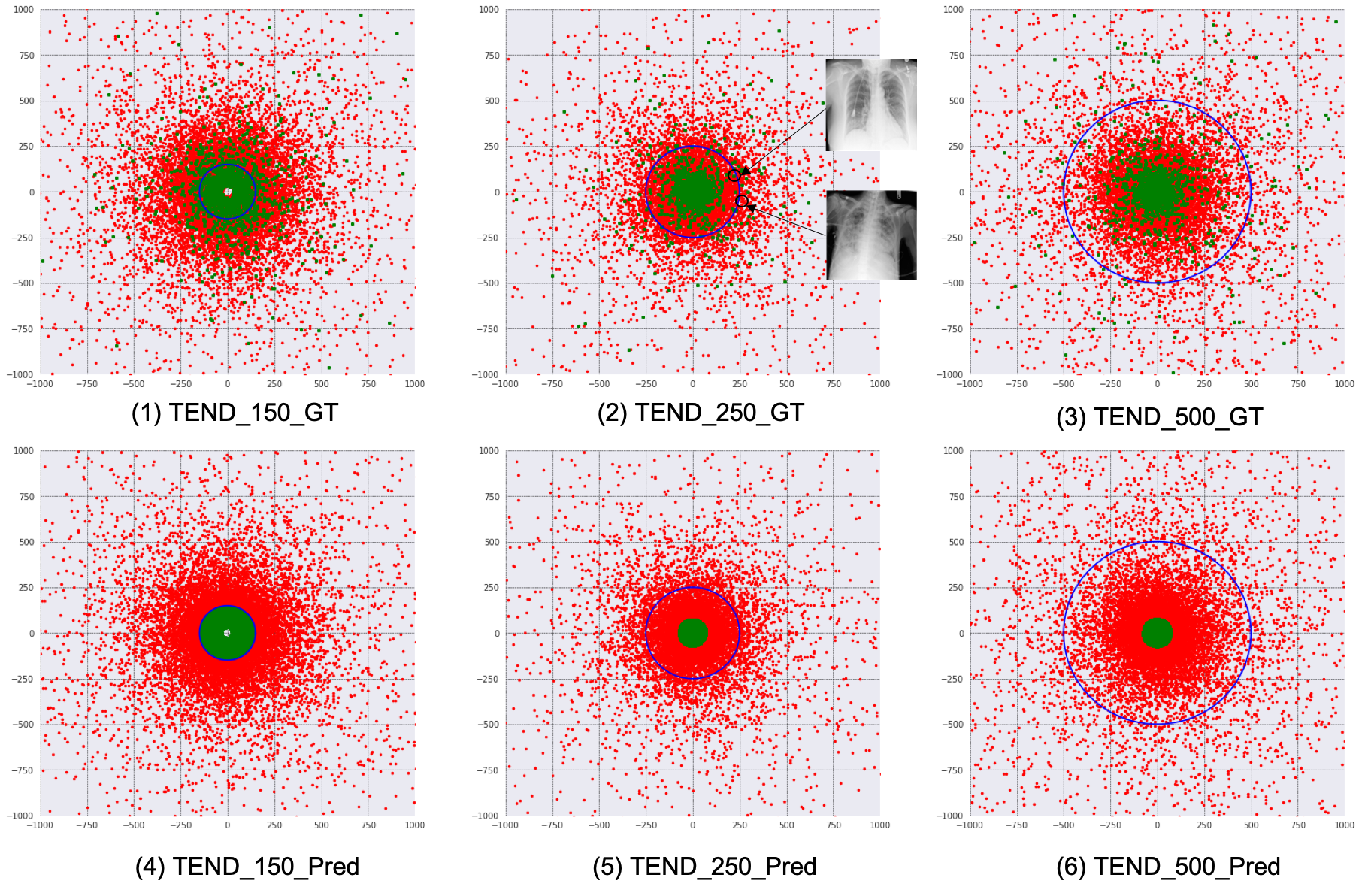}
\end{center}
  \caption{2D visualization of ID (green points) and OOD (red points) data distance distributions for RSNA dataset learnt by TEND's margin learner module with radius 150 (1st column), 250 (2nd column) and 500 (3rd column) \textbf{under the one-vs-rest setting}. The first row is for distance distribution with ground-truth labels; the second row shows the predicted results with the optimal threshold values. Blue circles are plotted based on the radius in each subfigure for reference.} 
\label{fig:rsna}
\end{figure}

\noindent As our model TEND has a margin learner module (see the $L_{mrg}$ part of Fig.~\ref{fig:model}) to enforce ID data inside of a predefined margin $R$ (illustrated as the green dotted circle in Fig.~\ref{fig:model}) as to the voted center $O$ (represented as the red star in Fig.~\ref{fig:model}) and OOD data outside of the region, we hereby visualize the data samples based on the obtained distance output by the \textit{MarginLearner}. Take one-vs-rest setup results for illustration, the voted center $O$, whose calculation details were introduced in Sec.~\ref{jointraining}, is located at the origin of the 2D coordinate system. To visualize each data sample, we utilize their distance to the voted center \textit{O} as their corresponding radius values to the origin. Each sample is represented by randomly picking one point along the circle that is defined with its corresponding radius. The x-axis and y-axis values help indicate how far the point is from the origin. Given an example with a distance value $d_{i}$, its corresponding coordinate ($x_{i}, y_{i}$) satisfies that $d_{i}^{2} = x_{i}^{2} + y_{i}^{2}$. The data samples with in-distribution labels are marked in green and the left data with OOD labels are in red. We draw the defined margin of the model with a blue circle for reference. Please refer to the Appendix code snippet for the visualization implementation details. Take RSNA dataset for example, in Fig.~\ref{fig:rsna}, the voted center $O$ is represented by the point with coordinates (0, 0) and the area defined by radius $R$ is present with the plotted blue circles in each subfigure. For better visualization and comparison, each subfigure has both the x-axis and y-axis ranging from -1000 to 1000, those data points that have larger distance out of range will be ignored. The first row shows the distance distribution of data with ground-truth labels (\textit{i.e., ID (in green) and OOD (in red)}) learnt by TEND with radius of 150 (1st column), 250 (2nd column) and 500 (3rd column), while the second row indicates the predictions after thresholding, with the green points for samples predicted as ID and red points for samples predicted as OOD. To help inspect the data points around the boundary, two cases based on the ground-truth information are illustrated for TEND\_250\_GT, with the upper one as an ID data and the lower case for OOD class.
From the first row, the learnt distance distributions for ID and OOD data are similar for TEND with different radius values. But the ID data can be outside the circle with radius 150 (subfigure (1)) but will be inside the circle regions with radius 250 (subfigure (2)) and 500 (subfigure (3)) of Fig.~\ref{fig:rsna}, which suggests that when using larger margin to divide ID and OOD data, ID samples will be easier to be included while more OOD data will be inside the region, leading to more false positive predictions. Therefore, it is not the larger the margin, the better the performance. 
After having the distance values predicted by the margin learner module, we apply the Gmeans method to find the optimal threshold considering both the distance predictions and the binary possibility. The second row illustrates the ID and OOD predictions of TEND after thresholding. We can see that the boundary of predicted ID data samples is very close to the margin circle of radius 150 (subfigure (4)), but much smaller compared to radius 250 (subfigure (5)) and 500 (subfigure (6)). As they are in the same scale, we can observe that the thresholding areas for ID are smaller when the margin values increase. 
\begin{figure}[htp]
\begin{center}
  \includegraphics[width=\linewidth]{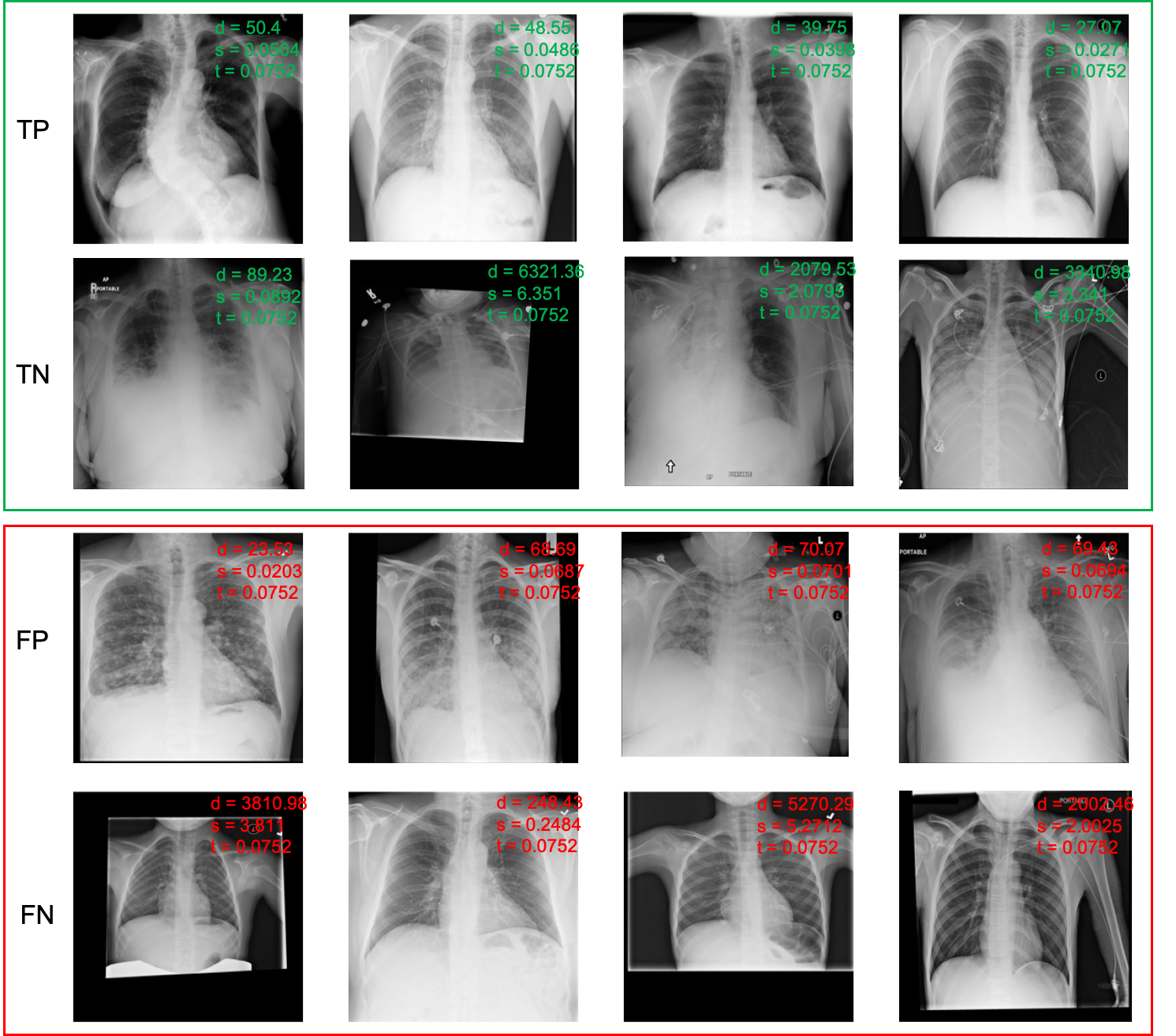}
\end{center}
  \caption{True Positive (TP, 1st row), True Negative (TN, 2nd row), False Positive (FP, 3rd row), and False Negative (FN 4th row) predictions of TEND\_500 on RSNA datasets \textbf{following the one-vs-rest setting}. d: distance value from the margin learner module, p: probability outputted by the binary discriminator module, s: final score, t: optimal threshold (ID: $s<t$, OOD: $s>=t$).} 
\label{fig:rsna_cases}
\end{figure}

To further analyze the OOD detection ability of TEND, we take the RSNA dataset for example and inspect part of the predictions. As shown in  
Fig.~\ref{fig:rsna_cases}, four kinds of predictions, namely true positive, true negative, false positive and false negative predictions, 
predicted by TEND\_500 are present, with four representative cases for each situation. TP means that true ID samples are correctly identified and TN is for correct identification of OOD samples. FP refers to the OOD data that is mis-classified as ID data and FN stands for wrongly classified OOD data.
From Fig.~\ref{fig:rsna}, data points close to the center are more confident of being in the ID category, which means the smaller the distance, the higher possibility of the data being an ID sample. Observing the TP cases in Fig.~\ref{fig:rsna_cases}, most of them are with distance values less than 50, which is relatively small compared to the predefined margin 500; while the TN cases are often with larger distances. The first chest X-ray image of TN cases has a final score 0.0892, close to the threshold 0.0752, which indicates this case is a challenging case. The third row of Fig.~\ref{fig:rsna_cases} are the hard FP cases for TEND\_500 to identify as they are all with both small distance values and probabilities. The FN cases shown in the fourth row of Fig.~\ref{fig:rsna_cases} can be those ID data with irregular format or position shifting. With imperfections, TEND\_500 will treat them as outliers and assign larger distance values by the margin learner module. Compared with others, the second FP case is much more challenging as the data is inside the predefined margin but classified wrongly due to the threshold setting. We also present the 2D distance visualization and detection results with examples for IVC-Filter and ISIC2019 datasets in the supplementary material.

\subsection{Effects of transformations}\label{nonlinear_trans}
To further compare the intra-class OOD detection ability, we generate validation data by applying four unseen transformations to all the ID data defined in Sec.~\ref{Transformations} and shown in the right yellow box in Fig.~\ref{fig:example}. As we have two experimental settings - the one-vs-rest and the rest-vs-one, we report them in Table~\ref{tab:val} and Table~\ref{tab:val0000} respectively. The best and the second best accuracy results are bolded and underlined respectively. As all the validation data are in OOD category, we calculate the OOD detection accuracy based on the optimal threshold $t$ determined in Table~\ref{tab:acc} (corresponding to Table~\ref{tab:val}) and Table~\ref{tab:acc000} (corresponding to Table~\ref{tab:val0000}) for each model and each dataset. Those data with score $s>=t$ are labeled as OOD (which are true negative samples, TN in short) and the data having score $s<t$ are classified as ID class (which are false positive samples, FP in short). Accordingly, the detection accuracy is formulated as $ACC_{val}= TN / (TN + FP)$.  

\subsubsection{One-vs-rest results of transformations}
Table~\ref{tab:val} shows the accuracy of detecting the generated validation OOD data with different models with the one-vs-rest experimental setting.
\begin{table*} [htp]
\caption{Accuracy of various OOD detection methods trained on IVC-Filter~\cite{ni2020deep:ivc}, RSNA~\cite{wang2017chestx} and ISIC2019~\cite{codella2018skin} \textbf{with the one-vs-rest setting}. Bold denotes the best results and * indicates the model is supervised. }
\label{tab:val}
\resizebox{\textwidth}{!}{%
\begin{tabular}{|l|cccc|cccc|cccc|cccc|}
\hline 
\multirow{2}{*}{Methods}   & \multicolumn{4}{c|}{IVC-filter} &\multicolumn{4}{c|}{RSNA} &\multicolumn{4}{c|}{ISIC2019}  \\ \cline{2-13} 
& \textit{\begin{tabular}[c]{@{}c@{}}Random\\Cut\end{tabular}}     & \textit{\begin{tabular}[c]{@{}c@{}}Random\\Crop\&Resize\end{tabular}}    & \textit{Noise}  & \textit{\begin{tabular}[c]{@{}c@{}}Gaussian\\Blur\end{tabular}}   

& \textit{\begin{tabular}[c]{@{}c@{}}Random\\Cut\end{tabular}}      & \textit{\begin{tabular}[c]{@{}c@{}}Random\\Crop\&Resize\end{tabular}}       & \textit{Noise}  & \textit{\begin{tabular}[c]{@{}c@{}}Gaussian\\Blur\end{tabular}}  

& \textit{\begin{tabular}[c]{@{}c@{}}Random\\Cut\end{tabular}}       & \textit{\begin{tabular}[c]{@{}c@{}}Random\\Crop\&Resize\end{tabular}}       & \textit{Noise} & \textit{\begin{tabular}[c]{@{}c@{}}Gaussian\\Blur\end{tabular}}  \\ \hline 

AutoEncoder~\cite{mcclelland1986parallel}      & {\textbf{1.000$\pm$0.000}}    & {$0.371\pm0.036$}    & {$0.988\pm0.007$}  & {$0.064\pm0.009$}   & {$0.001\pm0.000$}   & {$0.029\pm0.002$}  & {$0.422\pm0.004$}   & {$0.000\pm0.000$}  & {$0.252\pm0.004$} & {$0.581\pm0.005$} & {$0.428\pm0.004$} & {$0.187\pm0.002$}\\
AE\_GMM    & {$0.110\pm0.001$}     & {$0.151\pm0.000$}   & {$0.142\pm0.001$}  & {$0.142\pm0.001$}   & {$0.660\pm0.003$}   & {$0.023\pm0.001$}  & {$0.577\pm0.007$}   & {$0.402\pm0.007$} & {$0.055\pm0.001$} & {$0.028\pm0.001$} & {$0.086\pm0.002$} & {$0.087\pm0.002$} \\
VAE~\cite{an2015variational:an}      & {$0.013\pm0.006$}    & {$0.137\pm0.031$}   & {$0.020\pm0.013$} & {$0.008\pm0.007$}    & {$0.990\pm0.001$}   & {$0.288\pm0.004$}  & {$0.438\pm0.005$}   & {$0.424\pm0.005$}  & {$0.027\pm0.001$} & {$0.241\pm0.004$} & {$0.434\pm0.003$} & {$0.364\pm0.004$} \\
DeepSVDD~\cite{ruff2018deep:ruff}        & {\textbf{1.000$\pm$0.000}}     & {$0.735\pm0.039$}    & {$0.607\pm0.024$}  & {$0.044\pm0.018$}  & {$ 0.604\pm0.003$}     & {$0.120\pm0.005$}    & {$0.642\pm0.006$}  & {$0.455\pm0.005$}  & {$0.985\pm0.001$}    & {$0.740\pm0.003$}    & {$0.567\pm0.003$}  & {$0.190\pm0.004$} \\
GANomaly~\cite{akcay2018ganomaly:akcay}          & {\textbf{1.000$\pm$0.000}}    & {$0.792\pm0.017$}   & {$0.727\pm0.030$}  & {$0.690\pm0.031$}  & {$0.959\pm0.003$}   & {$0.910\pm0.003$}   & {$0.330\pm0.005$}  & {$0.313\pm0.005$}  &{$0.919\pm0.003$} & {$ 0.608\pm0.005$}& {$0.306\pm0.002$} & {$0.348\pm0.003$}\\
f-AnoGAN~\cite{schlegl2019f}  & {$0.888\pm0.024$}     & {$0.699\pm0.034$}    & {$0.583\pm0.035$}  & {$0.501\pm0.052$}  & {$0.726\pm0.005$} & {$0.729\pm0.007$}  & {$0.386\pm0.003$}  & {$0.413\pm0.005$}  & {$0.665\pm0.007$} & {$0.431\pm0.004$} & {$0.410\pm0.005$}& {$0.391\pm0.004$} \\
\hline
TEND\_150 (\textbf{ours})    & {$0.951\pm0.007$}        & {$0.988\pm0.006$}       & {$0.921\pm0.017$} & {\textbf{1.000$\pm$0.000}}  & {\textbf{1.000$\pm$0.000}}       & {\textbf{1.000$\pm$0.000}}      & {\textbf{1.000$\pm$0.000}}  & {\textbf{1.000$\pm$0.000}}   & {\textbf{1.000$\pm$0.000}}  & {\textbf{1.000$\pm$0.000}} & {\textbf{0.997$\pm$0.000}} & {\textbf{0.997$\pm$0.000}} \\
TEND\_250 (\textbf{ours})    & {\textbf{1.000$\pm$0.000}}       & {\textbf{1.000$\pm$0.000}}       & {\textbf{1.000$\pm$0.000}}  & {\textbf{1.000$\pm$0.000}}   & {\textbf{1.000$\pm$0.000}}      & {\textbf{1.000$\pm$0.000}}     & {\textbf{1.000$\pm$0.000}}   & {\textbf{1.000$\pm$0.000}}   & {$0.996\pm0.001$}  & {$0.942\pm0.003$} & {$0.799\pm0.005$} & {$0.741\pm0.005$} \\ 
TEND\_500 (\textbf{ours})    & {$0.752\pm0.026$}      & {$0.861\pm0.026$}       & {$0.797\pm0.029$}& {$0.984\pm0.008$} & {\textbf{1.000$\pm$0.000}}      & {\textbf{1.000$\pm$0.000}}       &{\textbf{1.000$\pm$0.000}}   & {\textbf{1.000$\pm$0.000}}   & {$0.950\pm0.002$} & {$0.976\pm0.001$} & {$0.905\pm0.002$} & {$0.905\pm0.003$} \\ 
\hline
BinaryClassifier*    & {$0.963\pm0.003$}     & {$0.963\pm0.005$}    & {$0.509\pm0.001$}  & {$0.899\pm0.001$}  & {$0.499\pm0.006$}      & {$0.680\pm0.003$}   & {$0.281\pm0.004$}  & {$0.215\pm0.004$}  &{$0.271\pm0.006$} & {$0.762\pm0.004$} & {$0.498\pm0.005$} & {$0.491\pm0.006$} \\ \hline
\end{tabular}%
}
\end{table*}
Among all the models present in Table~\ref{tab:val}, the AutoEncoder~\cite{mcclelland1986parallel}, VAE~\cite{an2015variational:an}, DeepSVDD~\cite{ruff2018deep:ruff}, GANomaly~\cite{akcay2018ganomaly:akcay}, f-AnoGAN~\cite{schlegl2019f} and our TENDs are all unsupervised methods, while the BinaryClassifier marked with an asterisk is a supervised model that is trained with both ID data and partial true OOD data. 
{\textit{Random Cut} is relatively easy to distinguish compared to other transformations as multiple methods including DeepSVDD, GANomaly and f-AnoGAN can detect most of them all for the three datasets. In contrast, the \textit{Random Crop and Resize}, \textit{Noise} and \textit{Gaussian Blur} transformations are much more difficult for them to handle. Nonetheless, TEND architectures with different margins nearly achieve all the best and the second best accuracy for the test datasets. In summary, although TEND is an unsupervised model, it can still obtain stronger intra-class OOD identification ability and even outperform other state-of-the-art models and the supervised model BinaryClassifer on both IVC-Filter and RSNA datasets. This advantage is due to the benefits of transformations during training. }

\begin{table*} [htp]
\caption{Accuracy of various OOD detection methods trained on IVC-Filter~\cite{ni2020deep:ivc}, RSNA~\cite{wang2017chestx} and ISIC2019~\cite{codella2018skin} \textbf{with the rest-vs-one setting}. Bold denotes the best results and * indicates the model is supervised. }
\label{tab:val0000}
\resizebox{\textwidth}{!}{%
\begin{tabular}{|l|cccc|cccc|cccc|cccc|}
\hline 
\multirow{2}{*}{Methods}   & \multicolumn{4}{c|}{IVC-filter} &\multicolumn{4}{c|}{RSNA} &\multicolumn{4}{c|}{ISIC2019}  \\ \cline{2-13} 
& \textit{\begin{tabular}[c]{@{}c@{}}Random\\Cut\end{tabular}}     & \textit{\begin{tabular}[c]{@{}c@{}}Random\\Crop\&Resize\end{tabular}}    & \textit{Noise}  & \textit{\begin{tabular}[c]{@{}c@{}}Gaussian\\Blur\end{tabular}}   

& \textit{\begin{tabular}[c]{@{}c@{}}Random\\Cut\end{tabular}}      & \textit{\begin{tabular}[c]{@{}c@{}}Random\\Crop\&Resize\end{tabular}}       & \textit{Noise}  & \textit{\begin{tabular}[c]{@{}c@{}}Gaussian\\Blur\end{tabular}}  

& \textit{\begin{tabular}[c]{@{}c@{}}Random\\Cut\end{tabular}}       & \textit{\begin{tabular}[c]{@{}c@{}}Random\\Crop\&Resize\end{tabular}}       & \textit{Noise} & \textit{\begin{tabular}[c]{@{}c@{}}Gaussian\\Blur\end{tabular}}  \\ \hline 

AutoEncoder~\cite{mcclelland1986parallel}    & {\textbf{1.000$\pm$0.000}}    &  {$0.116\pm0.009$}    & {$0.627\pm0.014$} & {$0.032\pm0.005$}   & {$0.999\pm0.003$}   & {$0.705\pm0.004$}  & {$0.901\pm0.002$}   & {$0.001\pm0.000$}  & {$0.782\pm0.004$} & {$0.250\pm0.005$} & {$0.388\pm0.004$} & {$0.368\pm0.004$} \\
AE\_GMM      & {$0.131\pm0.010$}    &  {$0.206\pm0.011$}    & {$0.212\pm0.010$}  & {$0.220\pm0.010$}  & {$0.361\pm0.003$}   & {$0.319\pm0.004$}  & {$0.383\pm0.005$}  & {$0.396\pm0.005$}  & {$0.067\pm0.002$} & {$0.054\pm0.002$} & {$0.158\pm0.003$} & {$0.157\pm0.003$} \\
VAE~\cite{an2015variational:an}      & {$0.036\pm0.002$}    & {$0.460\pm0.008$}   & {$0.476\pm0.011$}  & {$0.487\pm0.010$}    & {$0.188\pm0.002$}   & {$0.849\pm0.003$}  & {$0.603\pm0.005$}   & {$0.596\pm0.004$}  & {$0.174\pm0.003$} & {$0.627\pm0.006$} & {$0.555\pm0.007$} & {$0.544\pm0.007$} \\
DeepSVDD~\cite{ruff2018deep:ruff}        & {$0.858\pm0.011$}   & {$0.529\pm0.006$}   & {$0.495\pm0.008$} & {$0.496\pm0.008$}  & {$0.905\pm0.001$}     & {$0.415\pm0.004$}   &  {$0.494\pm0.004$}  & {$0.425\pm0.003$} & {$0.827\pm0.003$}  & {$0.294\pm0.004$}  & {$0.524\pm0.005$}  & {$0.541\pm0.005$} \\
GANomaly~\cite{akcay2018ganomaly:akcay}          & {$0.785\pm0.008$}    & {$0.583\pm0.009$}    & {$0.577\pm0.013$}  & {$0.629\pm0.009$}  & {$0.999\pm0.000$}  & {$0.682\pm0.003$}   & {$0.792\pm0.003$}  & {$0.238\pm0.005$}  & {$0.979\pm0.001$} & {$0.694\pm0.003$}& {$0.464\pm0.004$} & {$0.476\pm0.004$} \\
f-AnoGAN~\cite{schlegl2019f}  & {$0.934\pm0.008$}     & {$0.594\pm0.013$}    & {$0.361\pm0.014$}  & {$0.344\pm0.012$}  & {$0.380\pm0.004$}  & {$0.373\pm0.004$}  & {$0.716\pm0.003$}  & {$0.300\pm0.004$}  & {$0.989\pm0.001$} & {$0.825\pm0.002$} & {$0.460\pm0.005$} & {$0.464\pm0.006$} \\
\hline
TEND\_150 (\textbf{ours})    & {\textbf{1.000$\pm$0.000}} & {\textbf{1.000$\pm$0.000}}      & {\textbf{1.000$\pm$0.000}} & {\textbf{1.000$\pm$0.000}}   & {\textbf{1.000$\pm$0.000}}       & {\textbf{1.000$\pm$0.000}}     & {\textbf{1.000$\pm$0.000}}   & {\textbf{1.000$\pm$0.000}}   & {\textbf{1.000$\pm$0.000}} & {\textbf{1.000$\pm$0.000}} & {\textbf{1.000$\pm$0.000}} & {\textbf{1.000$\pm$0.000}} \\
TEND\_250 (\textbf{ours})   & {\textbf{1.000$\pm$0.000}}       & {\textbf{1.000$\pm$0.000}}     & {\textbf{1.000$\pm$0.000}} &  {\textbf{1.000$\pm$0.000}} & {\textbf{1.000$\pm$0.000}}     & {\textbf{1.000$\pm$0.000}}     & {\textbf{1.000$\pm$0.000}}  & {\textbf{1.000$\pm$0.000}}   & {$0.995\pm0.001$}  &  {\textbf{1.000$\pm$0.000}} & {$0.998\pm0.000$} & {$0.997\pm0.001$}\\ 
TEND\_500 (\textbf{ours})   & {\textbf{1.000$\pm$0.000}} & {$0.997\pm0.002$}      & {$0.999\pm0.001$} & {\textbf{1.000$\pm$0.000}}  & {\textbf{1.000$\pm$0.000}}    & {\textbf{1.000$\pm$0.000}}     & {\textbf{1.000$\pm$0.000}}   & {\textbf{1.000$\pm$0.000}} & {$0.984\pm0.001$} & {$0.941\pm0.002$} & {$0.902\pm0.004$} & {$0.760\pm0.002$} \\ 
\hline
BinaryClassifier*    & {$0.025\pm0.005$}    & {$0.796\pm0.010$}  & {$0.659\pm0.009$}  & {$0.644\pm0.012$}  & {$0.927\pm0.002$}      & {$0.972\pm0.001$}  & {$0.984\pm0.001$}  & {$0.816\pm0.003$}  & {$0.100\pm 0.003$} & {$0.849\pm0.003$} & {$0.470\pm0.003$} & {$0.477\pm0.003$} \\ \hline
\end{tabular}%
}
\end{table*}

\subsubsection{Rest-vs-one results of transformations}
Table 6 presents the accuracy of detecting the generated validation OOD data with different models following the rest-vs-one experimental setting. AutoEncoder partially retains its sensitivity in random cut and noise transformations for both IVC-Filter and RSNA datasets. In general, VAE shows little advantages in transformed OOD detection except for the noise and gaussian blur OOD detection for ISIC2019 dataset. DeepSVDD, GANomaly, f-AnoGAN occasionally show advanced performance for different situations. Comparatively, TENDs show more stable results in accurate detection of the transformed OOD data, especially for both IVC-Filter and RSNA datasets. This stability for such intra-class OOD detection benefits from the learning process of training with transformation.
\section{Discussion and Limitations}
We implement TEND with three different margins and show our results across various medical datasets under different settings. Although our models show competitive performance and surpass other methods under certain situations, the margin parameter has to be tuned for specific usages. Depending on the data complexity and variance across classes of a dataset, 250 is a good starting point. The ability of separation OOD from ID does not always improve as the margin increases due to the data complexity. For datasets with clear class variations, the margin can be set larger accordingly and vice versa. Besides, TEND utilizes transformation to generate fake OOD samples for discriminative learning. Due to the large amount of possibilities, this work only exploits a limited number of possible transformations.

\section{Conclusion}
%Conclusion
In this paper, we introduced an unsupervised novelty detector - TEND, which can detect intra-class OOD data for medical applications in an open-world environment. TEND is a two-stage anomaly detector with a vanilla AutoEncoder trained on in-distribution data in the first stage to serve as feature extractors in the second stage and two modules - a margin learner module and a binary discriminator module - jointly trained in the second stage for separating in-distribution inputs from the non-linearly transformed counterparts. With no OOD data used in training, TEND is able to learn nuances from intra-class variations in medical image analysis problems and provide a stepping stone for developing rare disease diagnosis models with no sample images. Extensive results with the one-vs-rest and rest-vs-one experimental settings on multiple public medical image datasets demonstrate the effectiveness of our model. More general evaluations on data with unseen transformations further evince our model's generalization ability and robustness. In summary, an efficient novelty detection method for medical images has been developed that can be applied to discover unknown classes with only predefined normal data. We plan to extend this work by integrating TEND into real time imaging pipelines for inference of medical imaging models.

\clearpage
\section{Appendices}
Below is the code for plotting the 2D visualization figure of the data samples according to obtained distances.

\begin{python}
import numpy as np
import matplotlib as mpl
import matplotlib.pyplot as plt
mpl.style.use('seaborn')

def generate_point(R):
    # given a radius R, generate the coordinates x, y 
    # of a random point in the cricle 
    theta = np.random.uniform(0, np.pi * 2)
    x = R * np.cos(theta)
    y = R * np.sin(theta)
    return x, y

def plot_point(anomaly_score, dist, threshold, R, K = 1000):    
    # anomaly_score: list of anomaly scores, (N,)
    # dist: list of distance values output by the MRG part of TEND, (N,)
    # threshold: anomaly score threshold, a float number, 
    # samples with anomaly score smaller than the threshold are classified as ID, 
    # equal to or greater than the threshold are in OOD category
    # K: float number, deciding the x-axis and y-axis range for showing data
    
    Xs = [], Ys = [], Xs2 = [], Ys2 = []
    for i in range(0, len(anomaly_score)):
        x, y = generate_point(dist[i])
        if anomaly_score[i] < threshold:
            Xs.append(x)
            Ys.append(y)
        else:
            Xs2.append(x)
            Ys2.append(y)
    fig = plt.figure(figsize=(8,8))
    plt.scatter(Xs, Ys, c ="green", linewidths = 2, marker ="s", s = 2)
    plt.scatter(Xs2, Ys2, c ="red", linewidths = 2, marker ="o", s = 2)
    plt.xlim([-k, k])
    plt.ylim([-k, k])
    theta = np.linspace( 0 , 2 * np.pi , 300 ) 
    plt.grid(color = 'black', linestyle = '--', linewidth = 0.5)  
    plt.plot(R * np.cos(theta), R * np.sin(theta), color='blue')
    return fig
\end{python}

\begin{figure}[htp]
\begin{center}
  \includegraphics[width=\linewidth]{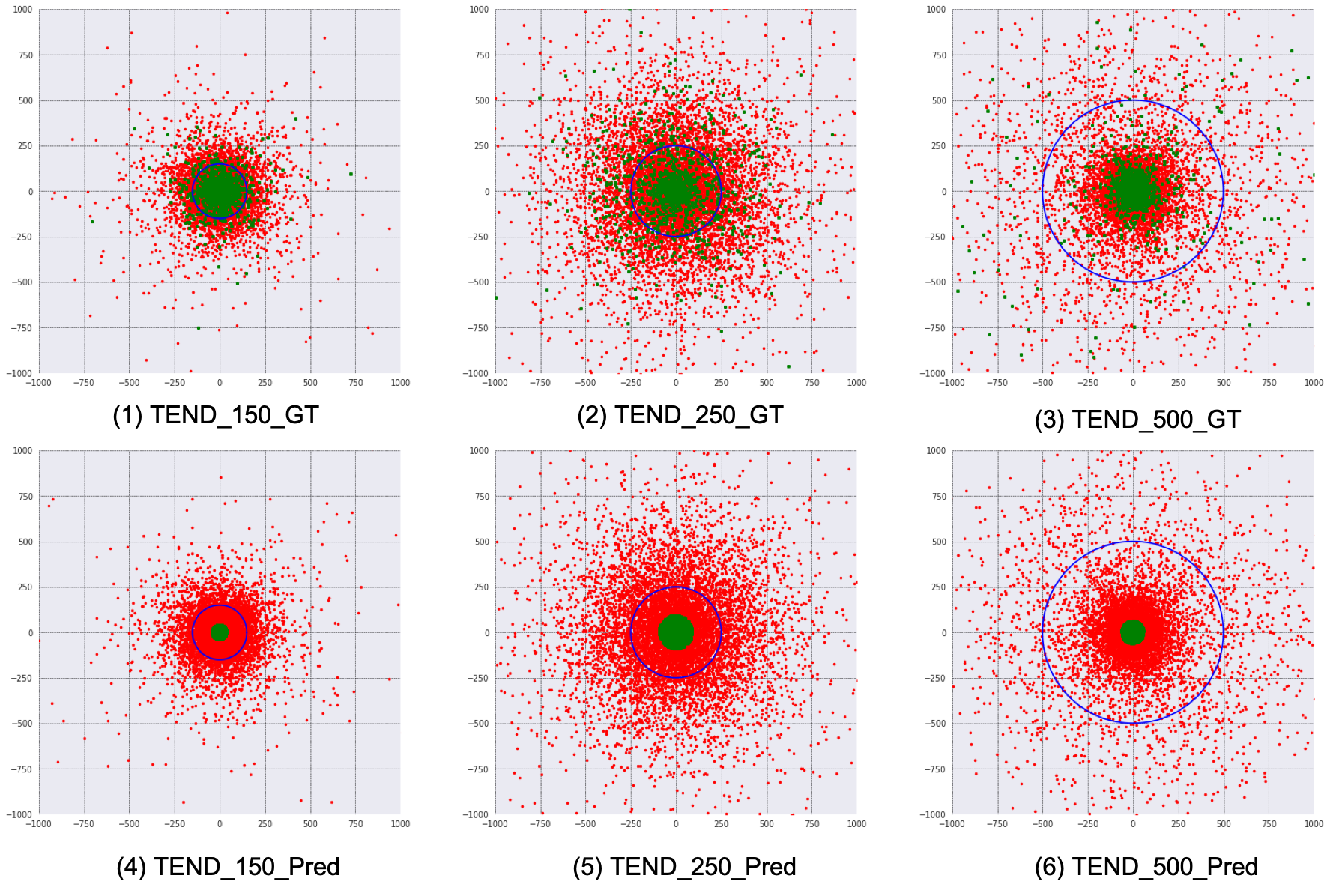}
\end{center}
  \caption{2D visualization of ID (green points) and OOD (red points) data distance distributions for ISIC2019 dataset learnt by TEND's margin learner module with radius 150 (1st column), 250 (2nd column) and 500 (3rd column) \textbf{following the one-vs-rest setting}. The first row is for distance distribution with ground-truth labels; the second row shows the predicted results with the optimal threshold values. Blue circles are the plotted based on the radius in each subfigure for reference.} 
\label{fig:isic}
\end{figure}

\begin{figure}[tp]
\begin{center}
  \includegraphics[width=\linewidth]{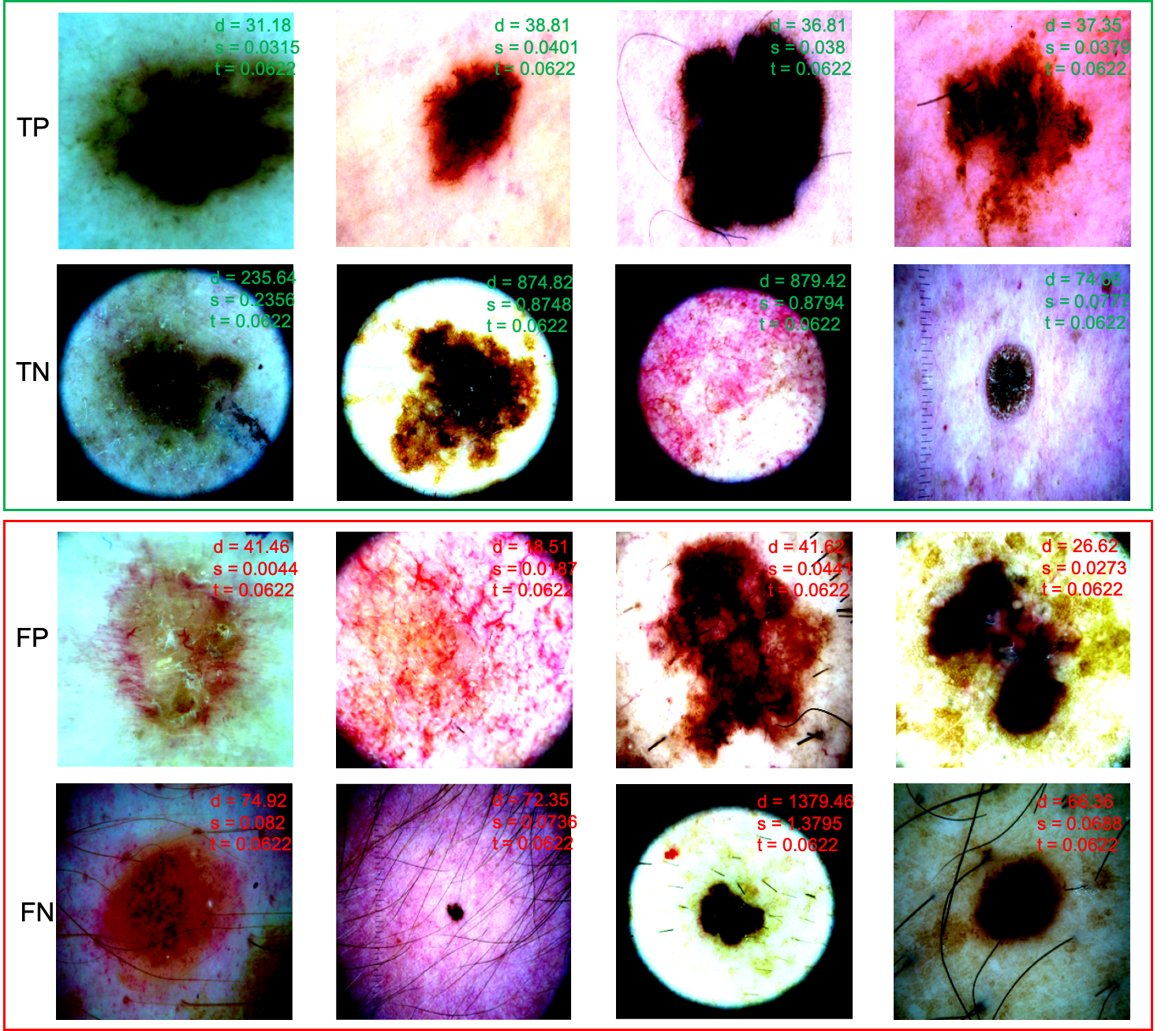}
\end{center}
  \caption{True Positive (TP, 1st row), True Negative (TN, 2nd row), False Positive (FP, 3rd row), and False Negative (FN 4th row) predictions of TEND\_500 on ISIC2019 datasets \textbf{with the one-vs-rest setting}. d: distance value from the margin learner module, p: probability outputted by the binary discriminator module, s: final score, t: optimal threshold (ID: $s<t$, OOD: $s>=t$).} 
\label{fig:isic_cases}
\end{figure}

\begin{figure}[tp]
\begin{center}
  \includegraphics[width=\linewidth]{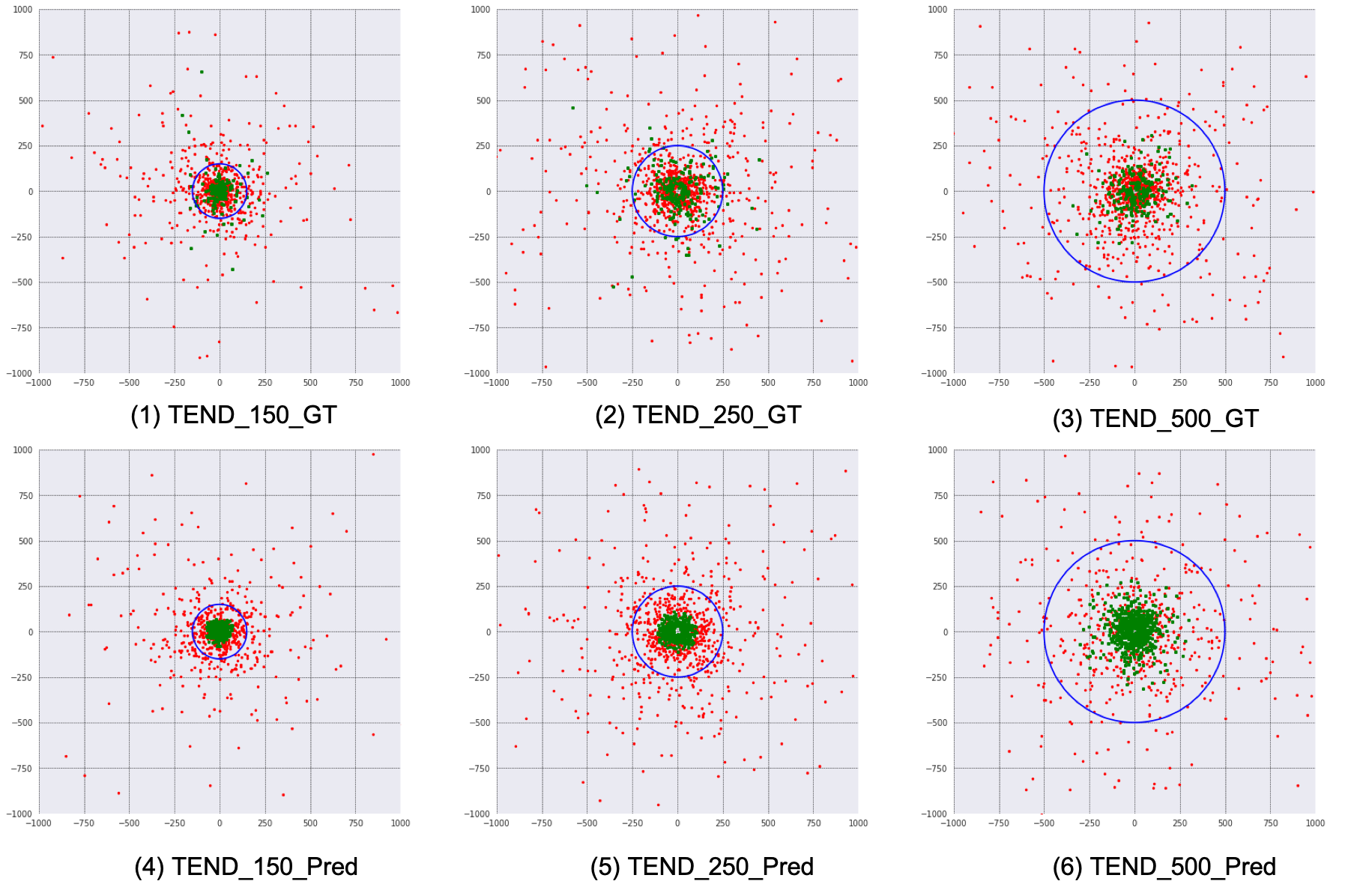}
\end{center}
  \caption{2D visualization of ID (green points) and OOD (red points) data distance distributions for IVC-Filter dataset learnt by TEND's margin learner module with radius 150 (1st column), 250 (2nd column) and 500 (3rd column) \textbf{under the one-vs-rest setting}. The first row is for distance distribution with ground-truth labels; the second row shows the predicted results with the optimal threshold values. Blue circles are the plotted based on the radius in each subfigure for reference.} 
\label{fig:ivc}
\end{figure}

\begin{figure}[tp]
\begin{center}
  \includegraphics[width=\linewidth]{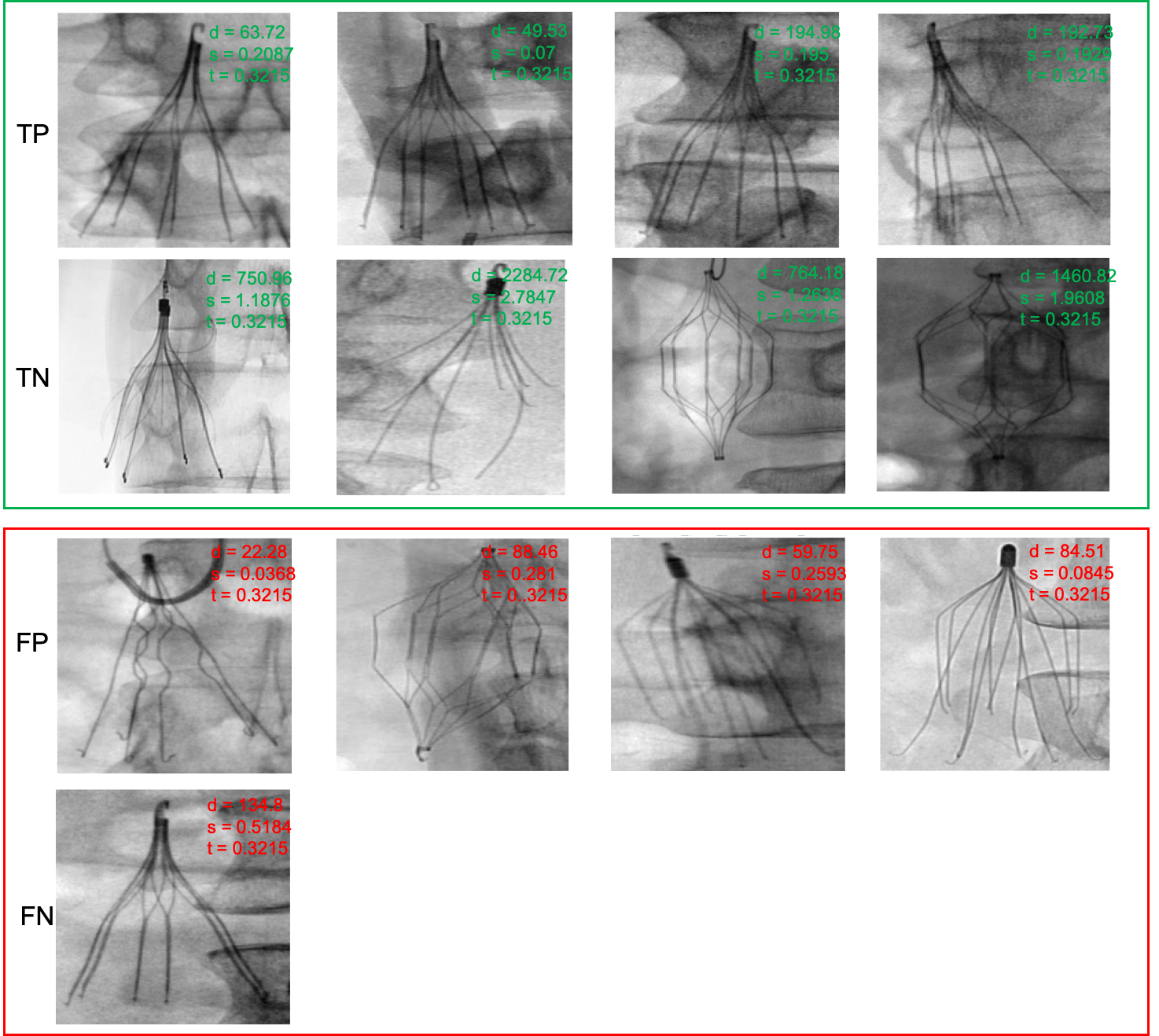}
\end{center}
  \caption{True Positive (TP, 1st row), True Negative (TN, 2nd row), False Positive (FP, 3rd row), and False Negative (FN 4th row) predictions of TEND\_500 on IVC-Filter datasets \textbf{following the one-vs-rest setting}. d: distance value from the margin learner module, p: probability outputted by the binary discriminator module, s: final score, t: optimal threshold (ID: $s<t$, OOD: $s>=t$).} 
\label{fig:ivc_cases}
\end{figure}

Here we show more 2D visualizations of ID and OOD data distance distribution for ISIC2019 dataset in Fig.~\ref{fig:isic} and  IVC-Filter dataset in Fig.~\ref{fig:ivc}. Different predictions including TP, TN, FP, FN are also present with the examples of RSNA dataset in Fig.~\ref{fig:isic_cases} and IVC-Filter dataset in Fig.~\ref{fig:ivc_cases}. Due to the limited False Negative predictions of IVC-Filter dataset, only one FP case is reported.

\clearpage

% \section{Disclosures}
% No conflicts of interests, financial or otherwise, are declared by the authors.

% \section{Acknowledgments}
% The work is supported by the National Institute of Biomedical Imaging and Bioengineering (NIBIB) MIDRC grant of the National Institutes of Health under contracts 75N92020C00008 and 75N92020C00021 and the US National Science Foundation \#1928481 from the Division of Electrical, Communication \& Cyber Systems.
%%%%% References %%%%%

\bibliography{article}   % bibliography data in report.bib
\bibliographystyle{spiejour}   % makes bibtex use spiejour.bst

%%%%% Biographies of authors %%%%%

% \vspace{2ex}\noindent\textbf{First Author} is an assistant professor at the University of Optical Engineering. He received his BS and MS degrees in physics from the University of Optics in 1985 and 1987, respectively, and his PhD degree in optics from the Institute of Technology in 1991.  He is the author of more than 50 journal papers and has written three book chapters. His current research interests include optical interconnects, holography, and optoelectronic systems. He is a member of SPIE.

\vspace{1ex}
% \noindent Biographies and photographs of the other authors are not available.

\listoffigures
\listoftables

\end{spacing}
\end{document}